\documentclass{IEEEtran}
\usepackage{blindtext, graphicx}
\usepackage{listings}
\usepackage{graphicx}
\usepackage[font=small]{caption}

\usepackage[utf8]{inputenc}
\usepackage{newunicodechar}
\newunicodechar{²}{\ensuremath{{}^2}}
\usepackage{float}
\usepackage{bbding}
\usepackage{pifont}
\usepackage{amssymb}
\usepackage{amsmath}
\usepackage{subfig}
\usepackage{graphicx}
\usepackage{float} 

\usepackage{color, colortbl}

\usepackage[switch]{lineno} 
\usepackage{lipsum} 

\usepackage{comment}
\usepackage{makecell}
\usepackage[]{acronym}
\newcommand{\xmark}{\ding{55}}%

\newcommand{\bx}{\mathbf{x}}

\newcommand{\by}{\mathbf{y}}
\newcommand{\bB}{\mathbf{B}}
\newcommand{\bH}{\mathbf{H}}
\newcommand{\bR}{\mathbf{R}}

\newcommand{\bM}{\mathbf{M}}
\newcommand{\bQ}{\mathbf{Q}}

\newcommand{\bz}{\mathbf{z}}

\definecolor{Gray}{gray}{0.85}

\title{Machine learning with data
assimilation and uncertainty quantification for dynamical systems: a review}

\author{
\IEEEauthorblockN{Sibo Cheng, César Quilodr\'an-Casas, Said Ouala, Alban Farchi, Che Liu, Pierre Tandeo,\\ Ronan Fablet, Didier Lucor, Bertrand Iooss, Julien Brajard, Dunhui Xiao, Tijana Janjic,\\ Weiping Ding,  Yike Guo, Alberto Carrassi, Marc Bocquet, Rossella Arcucci*}

\thanks{Sibo Cheng, César Quilodr\'an-Casas, Che Liu, Yike Guo and Rossella Arcucci are with Data Science Institute, Department of computing, Imperial College London, SW7 2AZ London, UK. }
\thanks{César Quilodr\'an-Casas, Che Liu and Rossella Arcucci are also with Department of Earth Science and Engineering, Imperial College London, SW7 2AZ London, UK. }
\thanks{Yike Guo is also with Department of Computer Science and Engineering, Hong Kong university of science and technology, Hong Kong, 999077, China. }
\thanks{Said Ouala, Pierre Tandeo and Ronan Fablet are with IMT Atlantique, Lab-STICC, UMR CNRS 6285,  France and  Odyssey, Inria/IMT, France.}
\thanks{Pierre Tandeo is also with RIKEN Center for Computational Science, Kobe, Japan}
\thanks{Alban Farchi and Marc Bocquet are with CEREA, \'Ecole des Ponts and EDF R\&D, \^{i}le-de-France, France.}
\thanks{Didier Lucor is with the Laboratoire Interdisciplinaire des Sciences du Num\'erique, CNRS, Paris-Saclay university, F-91403, Orsay, France. }
\thanks{Bertrand Iooss is with Electricité de France (EDF), 78401 Chatou, France, Institut de Mathématiques de Toulouse, 31062 Toulouse, France and SINCLAIR AI Lab, Saclay, France.}
\thanks{Julien Brajard is with Sorbonne University,  Paris, France and Nansen Environmental and
Remote Sensing Center (NERSC), Bergen, Norway.}
\thanks{Dunhui Xiao is with School of Mathematical Sciences, Tongji University, 200092 Shanghai, China.}
\thanks{Tijana Janjic is with Mathematical institute for machine learning and data science, KU Eichstätt-Ingolstadt, Bavaria, Germany.}
\thanks{Weiping Ding is with School of Information Science and Technology, Nantong University, 226019 Nantong, China. }
\thanks{Alberto Carrassi is with Department of Physics and Astronomy ``Augusto Righi'', University of Bologna, 40124 Bologna, Italy.}
\thanks{Corresponding author: Rossella Arcucci (r.arcucci@imperial.ac.uk)}
}

\begin{document}

\maketitle
\thispagestyle{empty}
\pagestyle{empty}

\begin{abstract}

Data Assimilation (DA) and Uncertainty quantification (UQ) are extensively used in analysing and reducing error propagation in high-dimensional spatial-temporal dynamics. Typical applications span from computational fluid dynamics (CFD) to geoscience and climate systems. Recently, much effort has been given in combining DA, UQ and machine learning (ML) techniques. These research efforts seek to address some critical challenges in high-dimensional dynamical systems, including but not limited to dynamical system identification, reduced order surrogate modelling, error covariance specification and model error correction. 
A large number of developed techniques and methodologies  exhibit a broad applicability across numerous domains, resulting in the necessity for a comprehensive guide. 
This paper provides the first overview of the state-of-the-art researches in this interdisciplinary field, covering a wide range of applications. This review aims at ML scientists who attempt to apply DA and UQ techniques to improve the accuracy and the interpretability of their models, but also at DA and UQ experts who intend to integrate cutting-edge ML approaches to their systems. Therefore, this article has a special focus on how ML methods can overcome the existing limits of DA and UQ, and vice versa. Some exciting perspectives of this rapidly developing research field are also discussed.


\end{abstract}

\section{Introduction}

The rapid growth of \ac{ML} has been witnessed in a wide range of research fields, including computer vision~\cite{voulodimos2018deep}, natural language processing~\cite{young2018recent} and AI for science~\cite{karniadakis2021physics}.
In particular, literature shows that the application of \ac{ML} algorithms, from conventional methods to deep neural networks, is present in nearly all aspects of spatio-temporal problems~\cite{ravuri2021skilful,dramsch202070,brunton2022data}. Adopting ML algorithms yields considerable improvements in forecasting complex high-dimensional dynamics. However, the black-box nature of ML algorithms makes them to exhibit poor interpretability, lack of robustness, weak reliability, and vulnerability to adversarial attacks and noisy systems.
On the other hand, \ac{DA}~\cite{carrassi2018data} and \ac{UQ}~\cite{abdar2021review} are reference frameworks that deal with model/data noises and error propagation inside dynamical systems. Compared to \ac{ML}, they provide interpretable and explicit solutions based on some mathematical assumptions, such as linearity and Gaussianity~\cite{carrassi2018data,smith2014,sullivan2015}.   

As~\cite{Geer2020} pointed out, substantial mathematical similarities exist between \ac{ML} and \ac{DA}, in particular, variational-type assimilation methods~\cite{carrassi2018data}. In fact, the latter also relies on gradient descent techniques to minimise a cost function measuring the difference between model outputs and prior estimation/observation. 
Several works have examined the connections between the acquisition, interpretation, and use of data in \ac{ML} and \ac{DA}. Integrations of \ac{DA} and \ac{ML} have been introduced in~\cite{brajard2020combining,arcucci2021deep,buizza2022data}. The link between probabilistic \ac{ML} approaches and differential equations is highlighted when the frameworks of \ac{DA} and \ac{ML} are combined from a Bayesian perspective. This equivalency, which demonstrates the parallels between the two areas, is presented formally in~\cite{Geer2020,bocquet2020bayesian}. Here, they show how to approximate Bayesian inverse methods (i.e., \ac{VarDA} in \ac{DA} and back-propagation in \ac{ML}) can be utilised to combine the four-dimensional \ac{VarDA} (4D-Var) and \ac{RNN} fields. In~\cite{bonavita2014data}, \ac{VarDA} and \ac{ML} are considered already incorporated in a Weak Constraint \ac{VarDA}, offering a somewhat different viewpoint. As demonstrated in~\cite{Raissi2016, bocquet2020bayesian,Perdikaris2017}, these approaches are also particularly well adapted to systems using Gaussian processes. These are data-driven algorithms capable of estimating model statistics and learning nonlinear, space-dependent, cross-correlations in a unified manner. Specifically, for high-dimensional systems, \ac{DA} is often combined with \ac{ROM}, such as classical \ac{POD}~\cite{berkooz1993proper} and \ac{ML}-based autoencoders~\cite{hinton2006reducing} to reduce the computational cost.
Both practical uses of these fusion algorithms, such as air quality forecasting using data-driven artificial intelligence~\cite{Lin2019, quilodran2021adversarial} and more theoretical ones, like spatiotemporal oscillations of the \ac{PDE}~\cite{Williams2014} using numerically computed approximations of Koopman eigenfunctions and eigenvalues, have been presented. Other methods, such as those in~\cite{brajard2020combining, quilodran2018fast}, which iteratively apply an \ac{EnKF} and a neural network to imitate hidden dynamics and forecast future states, are more akin to the works reported in this paper. A modular approach integrating neural network and \ac{DA} has been presented in~\cite{buizza2022data} which shows several methods to combine neural network and \ac{DA} to overcome limitations in applying these fields to real-world data. 

\begin{figure*}[h]
\centering
\includegraphics[width=0.9\textwidth]{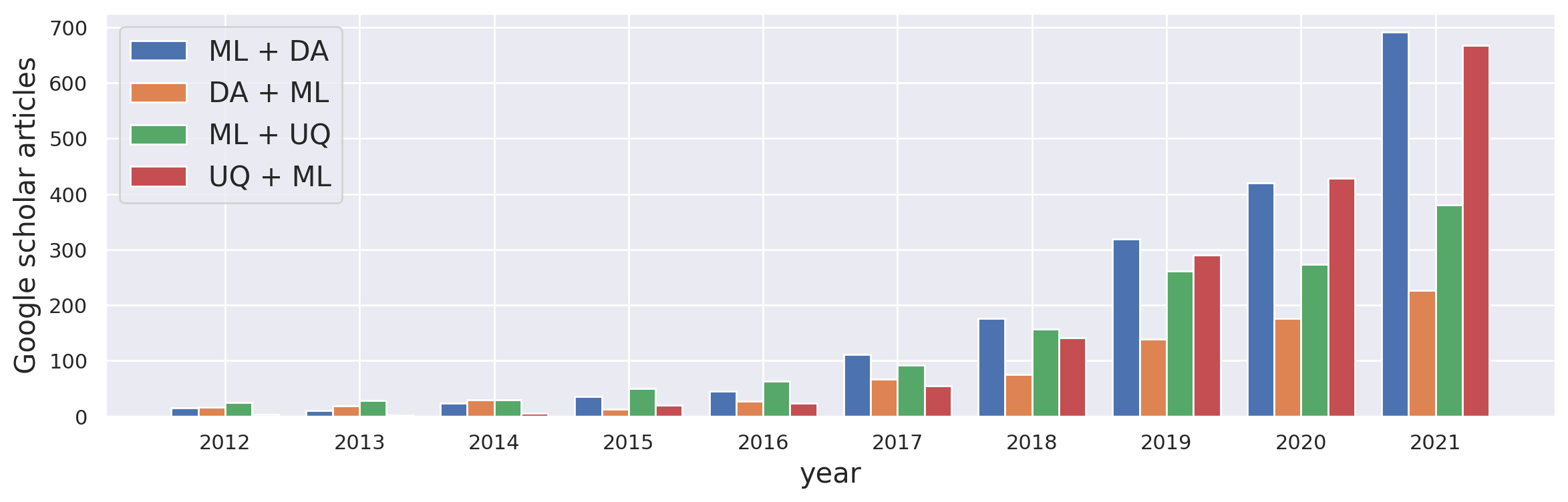}
\caption{Number of published articles combining \ac{ML}, \ac{DA} and \ac{UQ} according to Google scholar. 'A + B' denotes the number of articles which include 'A' in the title and 'B' in the text. }\label{fig:scholar}
\end{figure*}

\ac{ML} methods fail under their primary form in providing any guarantees of convergence or quantifying the error/uncertainty associated with their predictions, thus it is critical to provide \ac{UQ} to \ac{ML} predictions in order to anticipate and explain model failure to generalise. Model \ac{UQ} is crucial for instance to help choosing what data to learn from, or exploring an agent’s environment efficiently. In reality, data collection can be very expensive and time-consuming in dynamical system; and in extreme cases, only sparse and discrete batches of noisy data can be observed overtime. In these cases, \ac{UQ} for \ac{ML} helps in learning from small amounts of labelled data. 
\ac{UQ} is also extensively used for quantifying error propagation in dynamical systems~\cite{mezic2008uncertainty} through Monte Carlo methods and Polynomial Chaos~\cite{ghanem1999propagation,lucor2004generalized}. Monte Carlo methods are known for their broad applicability, and Polynomial Chaos is proven to have significant advantages in terms of computational efficiency and interpretability~\cite{mezic2008uncertainty}. When dealing with noisy dynamical systems, \ac{DA} and \ac{UQ} can be naturally combined. For example,~\cite{li2009generalized} made use of polynomial chaos expansion to model and reduce the sampling errors in \ac{EnKF}. A number of papers~\cite{borovikov2005multivariate,bousserez2015improved} applied Monte Carlo methods to estimate the error covariance matrices, which played a pivotal role in \ac{DA} algorithms.

Growing research efforts were devoted to combining and comparing \ac{DA} and \ac{UQ} with \ac{ML} under different contexts.
The number of published articles (including preprints on open-access repositories) from 2012 to 2021 that involved the concept of \ac{DA}, \ac{UQ} and \ac{ML} is illustrated in Figure~\ref{fig:scholar}. A sudden increase can be noticed, especially from 2015 when \ac{DL}~\cite{lecun_deep_2015} started to become the reference approach in many research areas. The applications of these methods cover a large range of fields, including climate science, fluid dynamics and image analysis. In this review, the related researches are mainly classified into two categories: \textit{\ac{DA} using \ac{ML} techniques} and \textit{\ac{ML} assisted by \ac{DA} and UQ}, respectively. The former focuses on \ac{ML}-based solutions to the long-standing challenges of \ac{DA}, including the correction of forward model errors and the error covariance specification. The second category gives attention to how \ac{DA} and \ac{UQ} can assist \ac{ML} in predicting high-dimensional dynamical systems. Specifically we concentrate on the challenge of noisy partial data and the use of real-time observations to progressively adjust \ac{ML} surrogate models.

\begin{figure}[h!]
\centering
\includegraphics[width=0.55\textwidth]{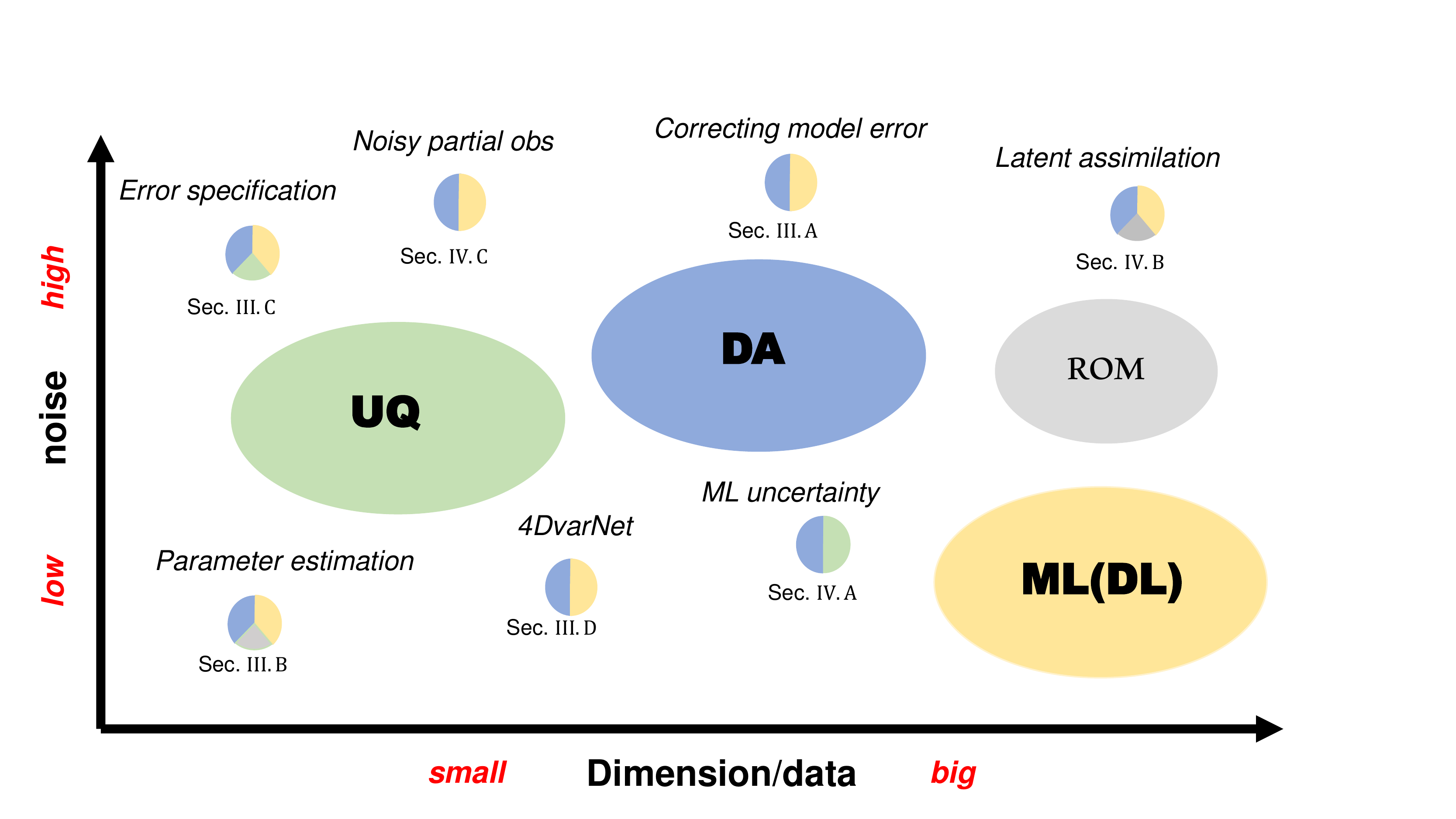}
\caption{Combination of \ac{ML}, \ac{DA} and \ac{UQ} methods and challenges versus available data dimension and noise level}\label{fig:star}
\end{figure}

Figure~\ref{fig:star} illustrates conceptually the related technologies as a function of problem dimension (x-axis) and noise level (y-axis) for their usual use cases. Different challenges presented in this review are also displayed where the colours indicate the technologies involved. It is worth mentioning that the \ac{ROM} plays a pivotal role in enabling the combination of \ac{ML} and \ac{DA} methods, especially in real-world applications, by reducing the computational cost. 

This review aims to cover most of the cutting-edge articles in the related research fields. This paper can thus serve as a comprehensive guide for navigating these fast growing techniques and methodologies.
We stress that the objective of this work is not to compare the performance of existing methods since they were developed to address different problems. In summary, we made the following contributions in this paper:
\begin{itemize}
    \item To the best of the authors' knowledge, this is the first review that addresses the combination of \ac{ML}, \ac{DA} and \ac{UQ} for dynamical systems.
    \item This paper has a special focus on how \ac{ML} methods can contribute to the key challenges of \ac{DA} and \ac{UQ}, and vice versa.  
    \item This review includes a range of main applications in \ac{DA}, \ac{UQ} and \ac{ML}, such as \ac{NWP}, environmental modelling and \ac{CFD}. 
    \item Some promising and insightful research perspectives and challenges are discussed.  
\end{itemize}

The rest of the paper is organized as follows. Section~\ref{sec:Preliminary} introduces the background and preliminaries for \ac{DA}, \ac{UQ} and \ac{ML} applied to high-dimensional dynamical systems. In Section~\ref{sec:DA use ML} and~\ref{sec:ML use DA}, we describe how the cutting-edge \ac{ML} techniques can be used to address the key challenges in \ac{DA} and \ac{UQ}, and vice versa. Other approaches and perspectives that combine \ac{ML} with \ac{DA} or \ac{UQ} are discussed in Section~\ref{sec: V}. We finish this review with a conclusion in Section~\ref{sec:conclusion}.

\section{Background and preliminaries}
\label{sec:Preliminary}
In this section, we briefly summarise the foundation of \ac{UQ}, \ac{DA} and \ac{ML} with a particular attention given to the application on high-dimensional dynamical systems. 

\subsection{Uncertainty and error quantification for complex and dynamical systems}
\label{sec:II-A}
In statistical terms, there are different types of uncertainties \cite{gal_uncertainty_2016,hullermeier_aleatoric_2021}, including
\begin{itemize}
    \item \emph{Aleatoric uncertainty} which originates in noisy input data (gappy, noisy, discordant or multimodal), where homoscedastic uncertainty denotes the variance that stays constant for all input parameters;
    \item \emph{Heteroscedastic uncertainty} represents the variance that depends on the input parameters and can potentially be predicted as a model output. In general, regardless of the quality of a model or the amount of training data, this uncertainty is irreducible;
    \item \emph{Uncertainty in model parameters} that best explain the observed data, for instance a large number of models are able to explain a given dataset, in which case we might be uncertain which model parameters to choose to predict with;
    \item \emph{Structural uncertainty}, i.e., what model structure should we use, how do we specify our model to interpolate and extrapolate well.
\end{itemize}

 The latter two can be grouped under model uncertainty that are epistemic uncertainties. Epistemic uncertainties describe the fidelity of the model in its representation of the data—barring aleatoric uncertainties. Typically in ML, epistemic uncertainties decrease as the training data size increases. Combining aleatoric and epistemic uncertainties provides us with the predictive uncertainty that is the confidence level of the model accounting for noise it can explain and noise it cannot. \\
Sensitivity analysis (SA) is the primary tool in UQ to  analyse error propagation in complex and dynamical systems through understanding and distinguishing the effects of various uncertainties on model output~\cite{saltar04,ghahig17,davgam21,rihan2003sensitivity}. SA can be used to determine which input variables contribute most to an output behavior, and which inputs are not influential, or to check for certain interaction effects in the model. This can be extremely useful when the model intrepretability is poor or the explicit formulas of the governing equation is out of reach, as in a majority of ML models.
The SA process involves calculating and analyzing sensitivity indices of input variables relative to a given quantity of interest in the model output (e.g., its mean, its variance, a particular quantile, its maximum). Considering the variance as an uncertainty metrics, sensitivity indices of each uncertain input variable on the variance of the model output allow for a better understanding of the model behavior, in order to reduce the uncertainties in the output in the most efficient manner. For example, identifying the most influential inputs will reduce their uncertainties, and then the model output uncertainty \cite{cacuci2005sensitivity}.

In \cite{davgam21}, four settings have been defined for the needs of SA in practice. First, the model exploration setting aims at understanding the behavior of the model by investigating the input-output relationship, e.g., via graphical tools. Second, the factors fixing setting aims at reducing the number of uncertain inputs by finding then fixing non-influential inputs. Third, the factors prioritization setting aims at precisely quantifying the effects of the most influential inputs. Last, the input distribution robustness setting aims at analyzing the variations in the quantity of interest with respect to uncertainty in inputs’ distributions.

Recently, it has been recognized (see, e.g., \cite{iooken22,ilibou22}) a wide analogy between SA and the topic of interpretability in ML \cite{mol22}.
As in SA, four settings have been defined in \cite{iooken22} for the needs of ML interpretability in practice: visualization of the relation between the predicted output label and the input features, identification of the most important features in ML prediction, important measures of explanatory variables and robustness of the decision boundary. 
\ac{UQ} can also be extremely useful in \ac{DA} models, especially in determining the prior and posterior errors of states and observations, which play an important role in \ac{DA}~\cite{d2013uncertainty}. 
In fact, as shown in the next section, \ac{DA} algorithms, particularly Kalman-based procedures, are probabilistic approaches where the state of the system is a Gaussian random variable. Thus, the state vector is defined at each time step by a mean vector and a covariance matrix. This \ac{UQ} of the assimilation results is important because it provides information on : the observation sampling in both space and time, the uncertainty correlation between variables, and the confidence in the estimate of the state vector.

\subsection{DA for dynamical systems}
\label{sec: DA principle}
\label{sec: II-B}
Data assimilation aims at predicting physical fields and estimating model parameters ~\cite{carrassi2018data, wang2000data} by aggregating information from different sources. Applying \ac{DA} to dynamical systems allows continuous corrections to model predictions while accounting for model and observation errors/uncertainties.

A typical time dependent \ac{DA} framework over a discrete time window $[0,...,T]$ involves variables and parameters as listed below.
\begin{itemize}
    \item states $\bx_t$: the target field of estimation, which is assumed unobservable in \ac{DA};
    \item true states $ \bx_t^\textrm{true}$: theoretical values of $\bx_t$;
    \item background states $ \bx_t^b$:  prior estimations of $\bx_t^\textrm{true}$, often obtained via predictive models;
    \item observations $\by_t$: observable quantities, for example, from sensors or satellites;
    \item transformation operators $\mathcal{H}_t$: functions that map the state variables to the observations;
    \item transformation operators $\mathcal{M}_t$: functions that map the dynamical system from $ \bx_{t-1}$ to $\bx_t$;
    \item prior errors $\epsilon_t^x, \epsilon_t^y$: estimation and prediction errors associated to $\bx_t^b$ and $\by_t$ respectively;
    \item error covariances $\bB_t, \bR_t, \bQ_t$: auto-covariance matrices of background, observation and model errors;
\item analysis states $\bx_t^a$: output of \ac{DA} models;
\end{itemize}
where $\by_t$ exists only when the observation at time $t$ is available. In the rest of this paper, we denote $\bx_{0:T} = \{\bx_0,...,\bx_T \}$ and $\by_{0:T} = \{\by_0,...,\by_T \}$ as a sequence of states and observations, respectively. We denote $\bH_t, \bM_t$ the linearisation of $\mathcal{H}_t, \mathcal{M}_t$.

Operational \ac{DA} models are mainly twofold: Kalman filter-based methods from estimation theory and variational \ac{DA} related to control theory. Both families of \ac{DA} approaches can be derived from Bayes' theorem~\cite{lorenc1986}.
The analysis states $\bx_t^a$ obtained from \ac{DA} could be viewed as a compromise between $\bx_t^b$ and $\by_t$, where the weights are determined by $\bB_t$, $\bQ_t$ and $ \bR_t$.
In \ac{DA}, the state vector is Gaussian because the errors terms $\epsilon_t^x$, $\epsilon_t^y$ in the state- and observation-space model are often assumed to be additive, Gaussian, and centered~\cite{carrassi2018data}. The background error represents the error propagation in the dynamical model and the fact that the initial condition is not necessarily well estimated~\cite{bannister2008review}. The observation error represents the mismatch between the observation vector ${\by}_t$ and the state vector projected in the observation space. This mismatch is mainly due to the error of representativity between the two vectors~\cite{janjic2018representation} and instrumental errors. Those background and observation errors are characterised by the covariance matrices ${\bB}_{t}$ and ${\bR}_{t}$, respectively. They show error amplitudes, error spatial correlations, and shared errors between variables.

\paragraph{Variational DA}
\label{sec:varDA}
 Following \ac{DA}'s statistical framework~\cite{lorenc1986}, we wish to directly apply Bayes' rule~\cite{lorenc1986} over  $[0, T]$, with batches of
observations $\mathbf{y}_{0:T}$. We can focus on the estimation of the conditional \ac{PDF} of $p(\mathbf{x}_{0:T}|\mathbf{y}_{0:T})$. Applying Bayes' rule, we obtain:
\begin{subequations} 
\begin{align}
  p(\mathbf{x}_{0:T}|\mathbf{y}_{0:T}) =& \frac{p(\mathbf{y}_{0:T}|\mathbf{x}_{0:T}) p(\mathbf{x}_{0:T})}{p(\mathbf{y}_{0:T})} \nonumber \\
  & \propto p(\mathbf{y}_{0:T}|\mathbf{x}_{0:T}) p(\mathbf{x}_{0:T}) ,
\end{align}
\end{subequations}
where the evidence $p(\mathbf{y}_{0:T})$ is inessential here.
We further assume that the observation errors are Gaussian and
uncorrelated in time, with covariance matrices $\bR_{0:T}$, so that:
\begin{subequations} 
\begin{align}
p(\mathbf{y}_{0:T}|& \mathbf{x}_{0:T}) = \prod_{t=0}^T p(\mathbf{y}_t|\mathbf{x}_t) \\
& \propto \exp\left[ -\frac{1}{2} \sum_{t=0}^T \|\mathbf{y}_t-\mathcal{H}_t(\mathbf{x}_t) \|^2_{\mathbf{R}_t^{-1}} \right].
\end{align}
\end{subequations}
Moreover, the prior \ac{PDF} $p(\mathbf{x}_{0:T})$ is assumed
\emph{Markovian}, i.e.~the state $\mathbf{x}_t$ conditional on the previous
state $\mathbf{x}_{t-1}$ does not depend on all other previous past
states:
\begin{subequations} 
\begin{align}
p(\mathbf{x}_{0:T}) =& p(\mathbf{x}_0)\prod_{t=1}^T p(\mathbf{x}_t | \mathbf{x}_{0:t-1}) \\
& \overset{\tiny \mbox{Markov}}{=} p(\mathbf{x}_0)\prod_{t=1}^T p(\mathbf{x}_t | \mathbf{x}_{t-1}) .
\end{align}
\end{subequations}
We also assume Gaussian statistics for the model error and the initial background which are uncorrelated in time, with zero bias and covariance matrices
$\bQ_{0:T}$ and $\bB_{0}$, respectively. Therefore, $p(\mathbf{x}_{0:T})$ is proportional to:
\begin{equation}
p(\mathbf{x}_0) \exp \left[
-\frac{1}{2} \sum_{t=1}^T \|\mathbf{x}_t- \mathcal{M}_t\left(\mathbf{x}_{t-1}\right) \|^2_{\mathbf{Q}_t^{-1}} \right],
\end{equation}
with 
\begin{equation}
    p(\mathbf{x}_0) = \textrm{exp} \left[
-\frac{1}{2} \|\mathbf{x}_0-\mathbf{x}_0^{b} \|^2_{\mathbf{B}_0^{-1}} \right].
\end{equation}

Now, we can gather the likelihood and prior pieces to obtain the cost function associated to the conditional \ac{PDF}
$p(\mathbf{x}_{0:T}|\mathbf{y}_{0:T})$:
\begin{subequations} 
\begin{align}
{\mathcal J}(\mathbf{x}_{0:T}) =&  -\ln \, p(\mathbf{x}_{0:T}|\mathbf{y}_{0:T}) \nonumber\\
=& \frac{1}{2} \Big( \|\mathbf{x}_0-\mathbf{x}_0^{b} \|^2_{\mathbf{B}_0^{-1}} +  \sum_{t=0}^T \|\mathbf{y}_t-\mathcal{H}_t(\mathbf{x}_t) \|^2_{\mathbf{R}_t^{-1}} \nonumber\\
&+  \sum_{t=1}^T \|\mathbf{x}_t- {\mathcal M}_t\left(\mathbf{x}_{t-1}\right) \|^2_{\mathbf{Q}_t^{-1}} \Big), \label{eq:4dvar}
\end{align}
\end{subequations}
up to constants that do not depend on the control variables $\mathbf{x}_{0:T}$.
This is the cost/loss function of the \emph{weak-constraint 4D-Var}~\cite{tremolet2006}, with which several connections to \ac{ML} can be made.
The cost/loss function can be minimised via gradient-based nonlinear optimisation, hence finding a compromise between the constraints of the observations, of the model, and of the background, whose errors are weighted against their statistics in the loss function.
The argument of the minimum is the analysis trajectory.
Note that to be applicable to geofluid models, this algorithm must be cycled in time, sequentially, i.e. the dynamical numerical model must be applied to this analysis.

Even though very powerful, these variational \ac{DA} methods require the tangent linear and adjoint models of ${\mathcal M}_{k}$ and ${\mathcal H}^{k}$~\cite{hascoet2014} which hampered their widespread adoption, but might now be boosted by the developments of differentiable models from \ac{ML}. More details are given in Section~\ref{sec: III-D} of this paper.

In variational \ac{DA}, well modelled error covariance matrices, including $\bB_t, \bR_t$ and $\bQ_t$, are required since they are crucial to spread information between observed and non-observed variables within the analysis of the scheme. However, due to the high-dimensionality, defining these covariances as a sequence of operators can be intricate and computationally demanding, as further discussed in Section~\ref{sec: III-C}.

\paragraph{Kalman-filter-based DA}
\label{sec:kalman}
Processing the measurements as they become available is what is done in a {\it filter}. As opposed to estimating the full \ac{PDF} $p(\mathbf{x}_{0:T}|\mathbf{y}_{0:T})$ all at once, filters sequentially estimate the marginal \acp{PDF} $p(\mathbf{x}_t | \mathbf{y}_{0:t})$, for all $ t \in [0,T]$. The process alternates an {\it analysis step}, based on Bayes' rule~\cite{gottwald2021combining}, where the conditional \ac{PDF} $p(\mathbf{x}_t | \mathbf{y}_{0:T})$ is updated using the latest observation $\mathbf{y}_t$, with a {\it forecast step} which propagates this \ac{PDF} to the next observation batch~\cite{jazwinski2007stochastic}.

This Bayesian approach is very difficult already in problems of moderate model dimension due to the cost of sampling and evolving these \acp{PDF}. Similar to the \ac{4DVar}, we assume that the uncertainties about observations, model, and prior are all Gaussian distributed: the \acp{PDF} are now defined by only means and covariance. By further assuming that the dynamical and observational models are both linear, with time-in-time and mutually uncorrelated errors, we get the \ac{KF}, which is the exact analytic solution to the Gaussian estimation problem:

\begin{subequations}

\begin{align}
\label{eq:KF-fcst1}
\mathrm{Forecast~Step} \qquad & \mathbf{x}^\mathrm{f}_t=\mathbf{M}_{t-1}\mathbf{x}^\mathrm{a}_{t-1}, \\
\label{eq:KF-fcst2}
& \mathbf{P}^\mathrm{f}_{t}=\mathbf{M}_{t-1}\mathbf{P}^\mathrm{a}_{t-1}\mathbf{M}_{t-1}^{\rm T}+\bQ_{t}.
\end{align}
\end{subequations}
\begin{subequations}
\begin{align}
\label{eq:KF-anl1}
\mathrm{Analysis~step} \qquad & \mathbf{K}_t=\mathbf{P}^\mathrm{f}_t \mathbf{H}_t^{\rm T}(\mathbf{H}^k\mathbf{P}_t^\mathrm{f}\mathbf{H}_t^{\rm T}+\mathbf{R}_t^{-1}), \\
\label{eq:KF-anl2}
& \mathbf{x}^\mathrm{a}_t=\mathbf{x}^\mathrm{f}_t+\mathbf{K}_t(\mathbf{y}_t-\mathbf{H}_t\mathbf{x}^\mathrm{f}_t), \\
\label{eq:KF-anl3}
& \mathbf{P}^\mathrm{a}_t= (\mathbf{I}-\mathbf{K}_t\mathbf{H}_t)\mathbf{P}^\mathrm{f}_t.
\end{align}
\end{subequations}
Equations~\eqref{eq:KF-fcst1}--\eqref{eq:KF-anl3} sequentially estimate the state, $\mathbf{x}^{\mathrm{f}}_t, \mathbf{x}^{\mathrm{a}}_t$, and error covariance, $\mathbf{P}^{\mathrm{f}}_t, \mathbf{P}^{\mathrm{f}}_a$. The matrix $\mathbf{K}_t$ 
is the Kalman gain containing the coefficients of the optimal linear combination between the prior mean and the observations. The analysis $\mathbf{x}^\mathrm{a}_t$, has minimum error variance and is unbiased.
The \ac{KF} is very powerful and, by solving for mean and covariance, provides a time dependent estimate of the system's state and associated uncertainty. Its biggest limitations are the linear assumptions and the computational cost for storing, evolving and manipulating the matrices. 

The extended Kalman filter (\ac{EKF},~\cite{jazwinski2007stochastic}) is a first-order expansion of the \ac{KF} for nonlinear dynamics. It operates a linearisation of the nonlinear model equations around the model's solution. The nonlinear model is used to propagate the state but the tangent linear model for the error covariance evolution. The \ac{EKF} also assumes Gaussian errors, but under the action of nonlinear dynamics, even an initial Gaussian error may become non Gaussian. The \ac{EKF} is therefore a good approximation as long as the observations interval is shorter than the time scale of the error growing modes~\cite{miller1994advanced}. Although the \ac{EKF} has been successful in a number of pioneering applications, including \ac{DA} for the geosciences, see {\it e.g.}~\cite{ghil1991data,de2014initialisation, evensen1992using}, it is also plagued by the same huge computational requirements as the \ac{KF}. 

A Monte Carlo approach is at the basis of a class of algorithms referred to as \acp{EnKF}~\cite{evensen2009data}. The \ac{EnKF}s use the \ac{KF} statistical framework and mimics its analysis updates, but the estimation and propagation of the errors is approximated by a finite ensemble of model realisations. While the accuracy of the \ac{EnKF}s is linked to the size of the computationally affordable ensemble, the \ac{EnKF}s gave proofs of extraordinary capabilities even in high-dimensional problems ($\mathcal{O}(10^8)$), by using ``only'' as few as $100$ members. The reasons behind this success are multiple and concurrent. The ensemble members are used to approximate a Gaussian distribution, as opposed to the vastly more complex task of estimating a generic, non parametric, \acp{PDF}. The latter is attempted by particle filters~\cite{van2019particle} that are in fact strongly affected by the curse of dimensionality. The application of the \ac{EnKF}s to chaotic dynamics, such as for geofluids, is challenged by the instabilities and the low predictability. Nevertheless, the \ac{EnKF}s benefit from the chaotic systems' tendency to confine error growths within a smaller subspace than the full system's dimension. Tracking this relatively low-dimensional unstable subspace with the finite ensemble is easier than affording the error description in the fully dimensional space~\cite{Carrassi2022}. 

Finally and in practice, the success of the \ac{EnKF}s in high dimensions is related to two ad-hoc fixes: \emph{inflation} and \emph{localisation} \cite{carrassi2018data}. Inflation consists in artificially increasing the ensemble-based error covariance, to combat error underestimation due to under sampling. Even more impactful, localisation acts to boost the ensemble-based error covariance rank and span, by reducing or even eliminating the small long distance correlations that are unavoidably poorly estimated with a small ensemble~\cite{asch2016}. A recent surveys on \ac{EnKF} from the mean field perspective and for both discrete and continuous time can be found in~\cite{calvello2022ensemble} while a review on state-of-the-art ensemble-based approaches in general, including ensemble smoother and ensemble variational methods, is given in~\cite{evensen2022data}.

\subsection{ML with UQ}
\label{sec:II-Bbis}

\begin{table*}[h]
\begin{center}

\caption{ML approaches considered in this review}%
\label{table:All_ML}
\begin{tabular}{cccc}
\hline
\hline
  Categories  & Methods & Application/section & References\\
\hline
\hline
& & & \\
\textit{Linear and polynomial}   & linear operator   & \ac{POD} (\ref{sec: II-C}), \ac{KF} (\ref{sec: II-B}), equation identification (\ref{sec: IV-C}) & \cite{Kalman1960,berkooz1993proper,9553130}  \\
  & SINDY   & ROM (\ref{sec: II-C}),  equation identification (\ref{sec: IV-C})& \cite{brunton2016discovering,fasel2022ensemble,brunton2016sparse}  \\
 & GLA (\ref{sec: IV-B})  & latent DA (\ref{sec: IV-B}), parameter estimation (\ref{sec: III-C})& \cite{Cheng2022JSC,cheng2022parameter,gong2022efficient}  \\
  & PCE   & UQ (\ref{sec:II-A} and~\ref{sec: IV-A})  & \cite{lucor2004generalized, li2009generalized}\\
  \hline
  & & & \\
 \textit{Neighborhood} & Kriging & equation identification (\ref{sec: IV-C})  & \cite{carnerero2022state} \\
& KNN & ROM (\ref{sec: II-C}) & \cite{cheng2022parameter,bai2021non,gong2022data}    \\
\hline
& & & \\
 \textit{Ensemble} & RF & ROM (\ref{sec: II-C}), parameter estimation (\ref{sec: III-C}) & \cite{cheng2022parameter,bai2021non}  \\
& DE & UQ (\ref{sec: IV-A}) & \cite{lakshminarayanan2017simple}    \\
\hline
& & & \\
 \textit{Deep learning} & CNN & ROM (\ref{sec: II-C}), error specification (\ref{sec: III-C}), latent DA (\ref{sec: IV-B}) & \cite{cheng2022data,liu2018deep,amendola2020,zhuang2022ensemble} \\
& RNN & ROM (\ref{sec: II-C}), error specification (\ref{sec: III-C}), latent DA (\ref{sec: IV-B}), equation identification (\ref{sec: IV-C}) & \cite{wang2016auto,cheng2022data,cheng2022observation,revach_kalmannet_2022}  \\
& BNN & parameter estimation (\ref{sec: III-C}), UQ (\ref{sec: IV-A}) & \cite{legler_janjic_2022,jospin2022hands}  \\
& GNN & ROM (\ref{sec: II-C}) & \cite{pfaff2020learning}  \\
& Transformer &  ROM (\ref{sec: II-C}), equation identification (\ref{sec: IV-C})& \cite{rakhimov2020latent,lin2022survey,fu2021data}  \\
 \hline
\hline
\end{tabular}

\end{center}
\end{table*}

Supervised and self-supervised \ac{ML} approaches are proven to be very efficient in many real-world applications and have still great potential for improving industrial means of production as well as research and development aspects. 
However, all these opportunities are still subject to methodological challenges as, among others, the bias-variance trade-off, the balance between the complexity of the underlying model to learn and the available amount of training data, a possible high dimensional input space, the presence of heterogeneous noise in the observations \cite{hastib09}.
One of these challenges concerns the UQ associated to \ac{ML} predictions.

ML techniques can be categorized in different families \cite{shaben14,dee20}, each of them having specific characteristics, for example, theoretical properties, practical performance in complex problems related to the amount of data it requires, efficiency in high dimension, stability, computational complexity and interpretability capabilities.
From the simplest to the most complex one, we distinguish the following four grand families:
\begin{itemize}
    \item Linear and polynomial models (into which the method of polynomial chaos fits \cite{legmar17}). Confidence intervals associated to predictions performed by these models can be obtained, as these models provide an analytical formula for leave-one-out error as well as mathematical properties of their regression coefficients \cite{bla09};
    \item Neighborhood models (into which the kriging method fits). The Gaussian process assumption behind the kriging model allows to associate easily computable confidence intervals associated to each prediction \cite{legmar17,demioo21}; 
    \item Ensemble models, especially those based on regression trees (e.g., random forests, gradient boosting). Obtaining confidence intervals for this kind of models is more difficult but two ways have been achieved: quantile regression \cite{mei09} and a specific subsampling procedure \cite{menhoo16};
    \item Deep learning, also known as Deep Neural Networks (DNN). In this category, different families of \ac{UQ} methods have been proposed based on Bayesian frameworks, for example, \ac{BNN}~\cite{jospin2022hands} and \ac{MCD}~\cite{gal2016dropout}. The latter consists of ensembles of \ac{NN} optimization iterates or independently trained \acp{NN} (e.g., deep ensembles: DE~\cite{lakshminarayanan2017simple}). Thorough overviews of uncertainty quantification in \ac{DL}, are provided by very recent review papers, e.g., \cite{gawlikowski_survey_2021,abdar2021review,psaros_uncertainty_2022}.

\end{itemize}
Finally, the method of conformal predictions \cite{shavov08} appears to be a valuable way to provide confidence intervals for any type of \ac{ML} models.

In the particular case of predicting high-dimensional dynamical systems, the state-of-the-art \ac{ML} methods often consist of \ac{DL}-based \ac{ROM} (see Section \ref{sec: II-C}). These reduced order models lie in a combination of different errors and uncertainties, including observation/data uncertainties, compression errors and predictive errors. The specification of these errors and the correction of the \ac{DL}-based \acp{ROM} are discussed in detail in Section~\ref{sec: III-C} and~\ref{sec: IV-A}, respectively. In this review, we consider the combination of \ac{DA} and \ac{UQ} with a wide range of \ac{ML} algorithms as shown in Table~\ref{table:All_ML}. However, the main focus is given to \ac{DL} approaches since they are state-of-the-art in predicting dynamical systems, especially in high-dimensional spaces.
\subsection{ML for predicting high-dimensional dynamical systems}
\label{sec: II-C}

\ac{ML} algorithms for predicting high-dimensional dynamics often rely on \ac{ROM}. More precisely, data are first compressed into a reduced latent space to decrease the computational cost. Predictive models are then used to surrogate the dynamics in the reduced space. Despite its efficiency, such a system will introduce several terms of errors and uncertainties, including compression and prediction errors. \ac{UQ} and \ac{DA} can be employed to specify and correct these errors, as discussed in detail in Section \ref{sec: IV-A} and \ref{sec: IV-B}.
In this section, we review the state-of-the-art \ac{ML} approaches for both \ac{ROM} and time-series predictions.

\subsubsection{Reduced-order-modelling}
\label{sec:ROM}

Reducing the dimension of complex dynamical systems has been a long-standing research problem~\cite{rega2005dimension,klus2018data}. Projection-based approaches, such as \ac{POD} and \ac{PGD} have been extensively applied in a large range of engineering problems~\cite{lucia2004reduced}, where the explicit transition from the full physical space to a low dimensional latent space relies on linear projection operators. In the past decade, much attention has been given in enhancing the \ac{ROM} using \ac{ML} methods, in particular, \ac{DL}-based autoencoders~\cite{wang2016auto}.
Autoencoder is a specific type of self-supervised neural network that has identical inputs and outputs. 
A typical autoencoder consists of an encoder $E$ which maps the input variables ${\bx}_t$ to the reduced latent space and a decoder $D$ which  reconstructs the full physical field ${\bx^\textrm{AE}_t}$ from the latent representation ${\bz}_t$, that is,
\begin{align}
    {\bz}_t = E(\bx_t) \quad \textrm{and} \quad {\bx^\textrm{AE}_t} = D({\bz}_t).
\end{align}
The encoder $E$ and the decoder $D$ are trained jointly with the objective to minimise the reconstruction loss, for instance, quantified by the \ac{MSE},
\begin{align}
    \mathcal{L}^{\textrm{MSE}} = \mathbb{E}\big( ||\bx_t - D \circ E (\bx_t)||_F^2 \big),
\end{align}
where $\mathcal{L}^{\textrm{MSE}}$ is the loss function. $||.||_F$ and $\circ$ denote the Frobenius norm and the composition function, repectively. $\mathbb{E}$ is the expectation operator.

 \begin{figure*}[h]
\centering
\includegraphics[width =0.9 \textwidth,height=10cm, trim={0cm 15cm 0 0}]{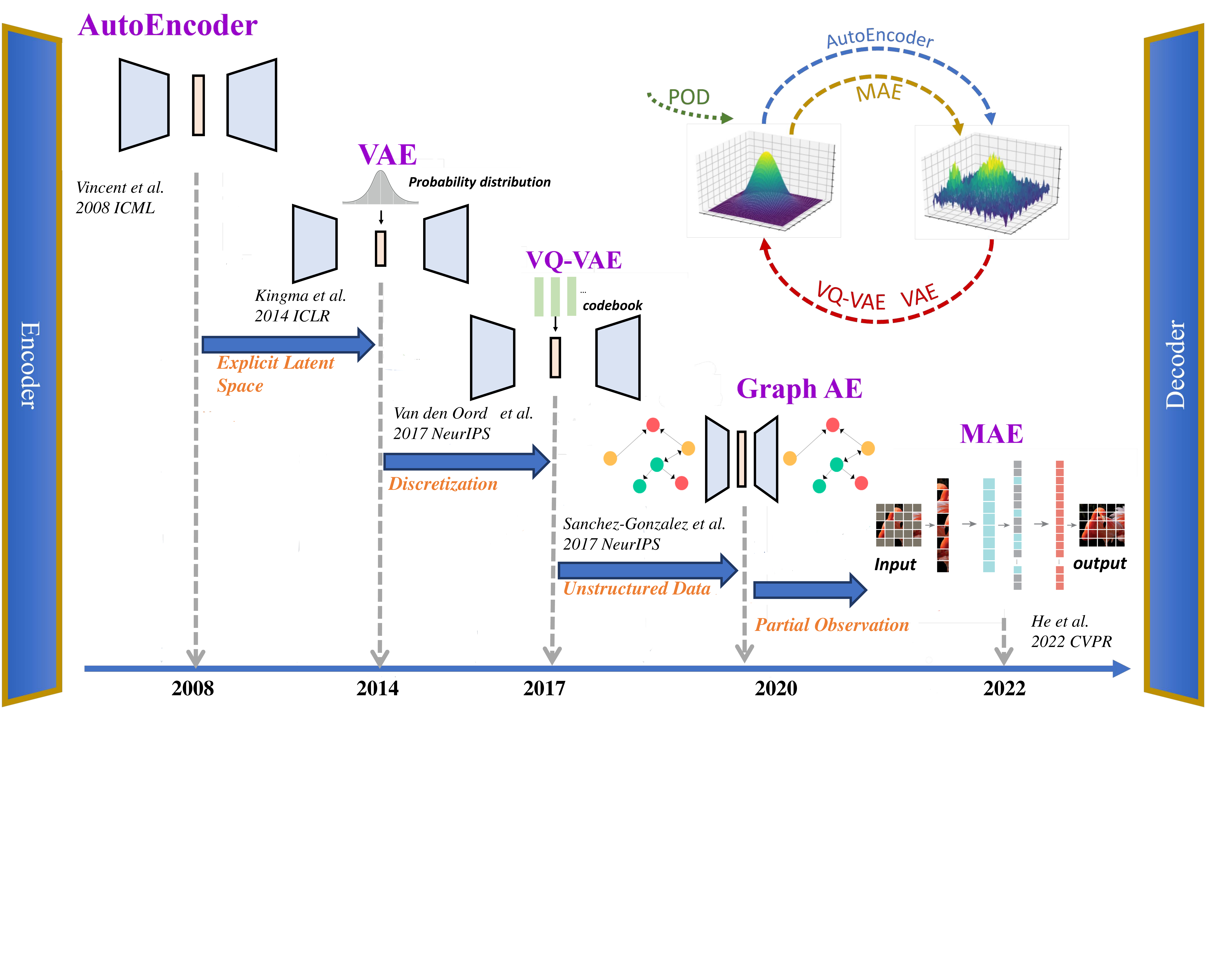}
\vspace{-2mm}
   \caption{Progression of \ac{ML}-based reduced-order-modelling}
   \label{fig:flowchart_inverse}
\end{figure*}

\acp{AE} show great potential in capturing nonlinear patterns compared to projection-based methods such as \ac{POD}~\cite{lee2020model, fu2021data, wu2021latent}. However, the geometry of the latent space obtained by \acp{AE} can be chaotic, and thus, less interpretable~\cite{mococxr}.
Continuous efforts have been given in combining \ac{DL} autoencoders with traditional dimension reduction methods. Carlberg et al.~\cite{carlberg2019recovering} used dimensionality reduction methods comprising \ac{PCA} and \acp{AE} to recover missing data. The works of~\cite{otto2019linearly} and~\cite{takeishi2017learning} used \ac{AE} to learn the Koopman invariant
subspace for \ac{DMD}. A number of studies have also successfully applied \ac{POD}-based \acp{AE} for urban air pollution~\cite{casas2020urban} and nuclear engineering~\cite{phillips2020autoencoder,gong2022efficient}. These methods benefit from both the accuracy of \ac{DL} \acp{AE} and the interpretability of projection-based approaches.
To tackle the issue of chaotic latent space, \ac{VAE} was proposed by~\cite{kingma2013auto}, where a regularisation term is added in the loss function. The latent variables were constrained by Gaussian distributions through the \ac{KLD} to ensure the smoothness of the latent space geometry~\cite{kingma2013auto}. The explicit latent space could naturally improve the interpretability of \acp{AE}. More recently, the work of~\cite{van2017neural} introduced Vector Quantized Variational Autoencoders (VQ-VAE) which generate a discrete latent space instead of a continuous one like standard \acp{VAE}. Some recent researches~\cite{polyak2019attention,fu2021data,song2019learning} also attempted to enhance the reconstruction performance by employing an attention-based mechanism~\cite{vaswani2017attention}. For example, Rui et al~\cite{fu2021data} presented a nonlinear non-intrusive using an \ac{AE} and self-attention \ac{DL} method. Stacked \ac{AE} is used to perform the nonlinear model reduction and a self-attention mechanism is used to represent the fluid dynamics.

Among different structures of \acp{AE}, \ac{CAE} (including convolutional \ac{VAE}), is by far the most widespread architecture.
However, it can be cumbersome to apply \ac{CAE} for unstructured data, for instance, in \ac{CFD} with irregular meshes.
To address this bottleneck, \ac{GNN} architectures~\cite{pfaff2020learning, sanchez2020learning} were proposed and applied with success for modelling liquids and granular materials.  Moreover,~\cite{pfaff2020learning} showed how graph-based \ac{ML} can also learn adaptive remeshing.
There is increased attention on using meshes for learned geometry and shape processing, but despite the widespread use in classical simulators, adaptive mesh representations have yet to see much use in learnable prediction models. To deal with incomplete data, the pioneering work of~\cite{he2022masked} introduced Masked autoencoders, capable of reconstructing the full field using a limited number of observable patches. 

The development of different \acp{AE} is illustrated in Figure~\ref{fig:flowchart_inverse} with a particular focus on the transitions between 'explicit' and 'implicit' latent spaces.

\subsubsection{Predictive models}
\label{sec:predictive}

Forecasts produced by \ac{ROM}s cost only a fraction compared to high-dimensional model solution. Non-intrusive \acp{ROM} were broadly used in predicting reduced variables. Traditional approaches often relied on, for instance, radial basis functions~\cite{xiao2016non} or shallow machine learning techniques, such as \ac{KNN} and \ac{RF}~\cite{cheng2022parameter}.  Recently, \acp{RNN} have been used to model and predict temporal dependencies between inputs and outputs of \ac{ROM}s. \ac{ROM}s and \ac{RNN}s are used together in previous studies, e.g.~\cite{quilodran2018fast, quilodran2021adversarially, reddy2019reduced} where the surrogate forecast systems can easily reproduce subsequent time-steps. \ac{LSTM} networks, originally described in~\cite{hochreiter1997long}, are used extensively to learn the underlying dynamics in the reduced space~\cite{nakamura2021convolutional,mohan2018deep}.
\ac{LSTM} is a special variant of \ac{RNN} that is stable and powerful enough to be able to model long-range time dependencies~\cite{xingjian2015convolutional} and overcomes the vanishing gradient problem~\cite{greff2016lstm}. Some recent works in \ac{ROM} also focused on the state-of-the-art \ac{ML} predictive models, namely Transformer~\cite{rakhimov2020latent} and adversarial predictions~\cite{cheng2020data, quilodran2021adversarial}. Transformers, originally developed in \ac{NLP}, have been successfully implemented with latent space representations of videos~\cite{rakhimov2020latent}, or time-series forecasting~\cite{tong2022probabilistic}.
However, the model efficiency and model adaptation are hampered during the implementations of transformers due to the computation and memory complexity of the self-attention modules in the transformer~\cite{lin2022survey}.
Some other approaches aim to learn the underlying governing equations with dynamical data as input. For example, \ac{SINDy}~\cite{brunton2016discovering, kaiser2018sparse} present a procedure that extracts sparse dynamic system models from time series data. \ac{SINDy} has been successful in generating robust, high quality models for physical systems, even with a \ac{ROM} obtained via \ac{PCA}~\cite{Kaptanoglu2022, paglia2022, cai2022online} or deep \ac{AE}~\cite{champion2019data}.
Conversely, the accuracy of predictions can also be increased by including specialised knowledge about the system modelled in the form of loss terms~\cite{kim2019deep, wiewel2019latent}, or by physics-informed feature normalisation~\cite{thuerey2020deep}.    
In summary, various \ac{ML} predictive models were paired with \ac{ROM} to release the computational burden in high-dimensional system modelling.    
However, when the predicted output is used as an input for the prediction of the subsequent time sequence (known as the 'rollout' process), the results can detach quickly from the underlying physical model solution when encountering out-of-distribution data. This detachment is mainly due to the error/uncertainty propagation and accumulation during iterative predictions over rollouts~\cite{sanchez2021learning}.

\section{Data assimilation using machine learning techniques}
\label{sec:DA use ML}
In this section, we focus on how \ac{ML} techniques are used to address the key challenges of \ac{DA} algorithms, including model error correction (Section~\ref{sec: III-A}), parameter estimation (Section~\ref{sec: III-B}), error covariance specification (Section~\ref{sec: III-C}) and end-to-end learning of \ac{DA} system (Section~\ref{sec: III-D}).  

\subsection{ML to correct model errors in DA}
\label{sec: III-A}

As discussed at the end of Section~\ref{sec:predictive}, when predicting complex dynamical systems, physics-based or data-driven numerical models are inevitably affected by errors.
Typical model error correction approaches consist of using a statistical model (typically a neural network) to correct a physical, knowledge-based model. In practice, this implies that we try to build a hybrid physical/statistical model~\cite{rasp-2018, bolton-2019, jia-2019, watson-2019, bonavita-2020, brajard2021combining, gagne-2020, wikner2021using, farchi2021using}. A physical model is usually defined by a set of differential equations. These equations are discretised and implemented to form the model tendencies. A numerical scheme is then used to integrate the tendencies over a small time interval and several integration steps are composed to iterate from one time to the next. From here, there exist various ways to design the hybrid model, depending on how the statistical correction is introduced~\cite{farchi-2021a}. The easiest possibility is to include a single correction per model integration:
\begin{equation}
\label{eq:hybrid}
    \mathbf{x}_{t+1}=\mathcal{M}_t\left(\mathbf{x}_{t}\right) + \mathcal{F}_t\left(\mathbf{x}_{t}\right),
\end{equation}
where $\mathcal{M}_t$ is the resolvent of the physical model from $t$ to $t+1$ and $\mathcal{F}_t$ is the statistical correction, written in an additive form for simplicity (there are of course other possibilities such as multiplicative correction). This is called {\it resolvent correction} because, in this case, the correction is added to the resolvent. At the other end of the spectrum, the statistical correction can be included in the model tendencies. The {\it tendencies correction} is potentially more efficient because the errors can be corrected before they manifest (i.e. before they are integrated) and because the statistical correction benefits from the interaction (via the integration scheme) with the physical model. However, a tendency correction is by construction intrusive (even and prominently at the level of physical and statistical models' codes interdependence) and hence more difficult to implement than a resolvent correction. This approach can be formalised in the  general form,
\begin{align}
  \mathbf{x}_{t+1}=\mathcal{F}_t\left(\mathcal{M}_t\left(\mathbf{x}_{t}\right), \mathbf{x}_{t}\right). 
\end{align}
 It enables representation of non-additive error~\cite{wikner2021using} or to increase the model resolution from the original resolution of the physical model to a higher dimension~\cite{barthelemy-2022}. On the other hand, the correction can no longer be directly related to the analysis increment.

\begin{figure}[h!]
    \centering
    \includegraphics[width=0.50\textwidth]{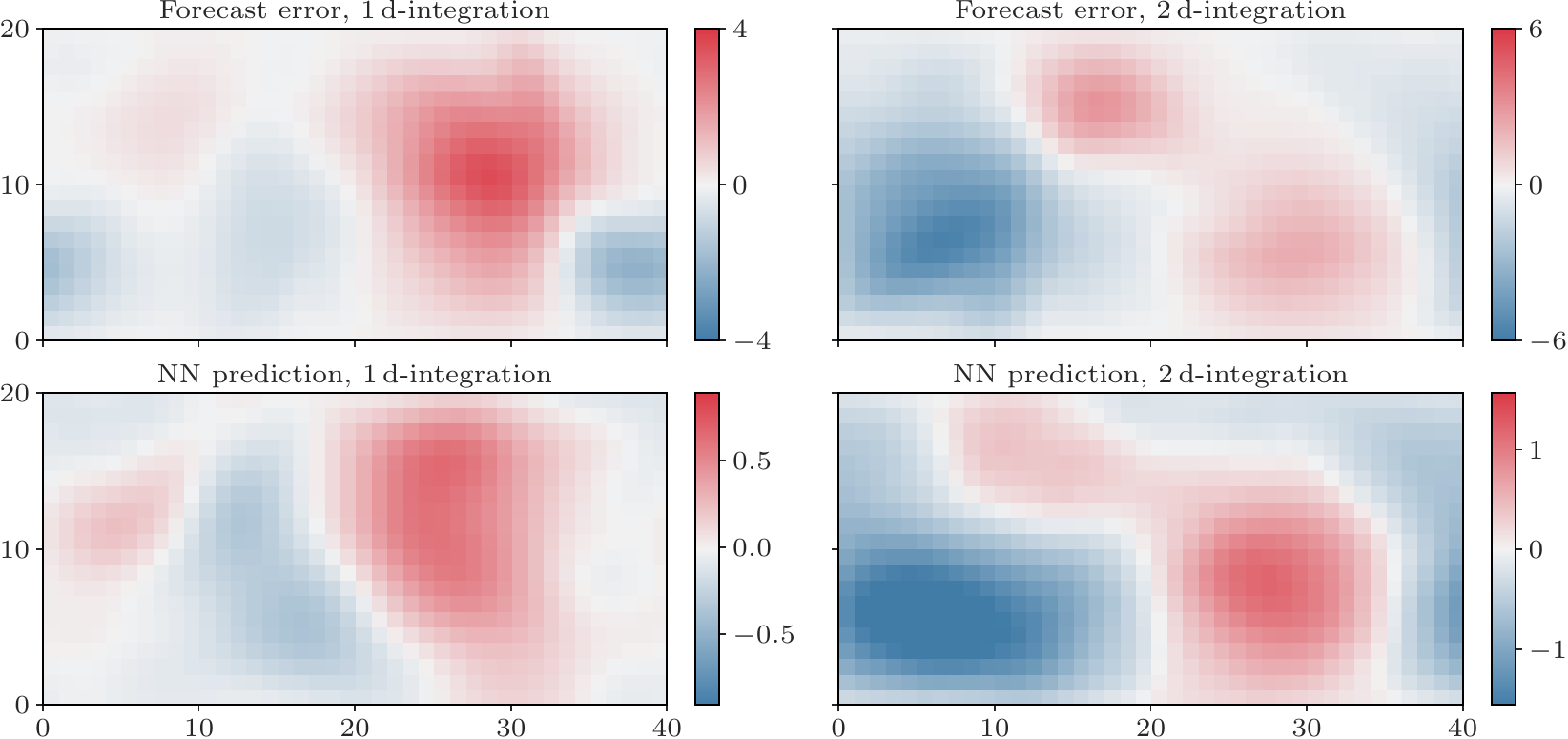}
    \caption{Correction of a two-dimensional quasi-geostrophic model using a neural network. Top panels: true model error snapshots for a 1-day (left) or 2-day (right) integration. Bottom panels: corresponding neural network predictions, based on the analysis increments. Figure reproduced from~\cite{farchi2021using}.}
    \label{fig:mec-snap}
\end{figure}

In a \ac{DA} context, where observations are usually sparse and noisy, the statistical correction can be trained using a series of \ac{DA} analyses, where the system state has first been estimated from observations~\cite{brajard2020combining}. In the case of the resolvent correction, the contribution of the physical and statistical models is independent. Therefore, it is possible to show that the statistical model can predict the analysis increments~\cite{farchi2021using, sacco2022evaluation}. This independence is convenient since the analysis increments are usually products of the \ac{NWP} centres. However, even though analysis increments can be seen as a proxy for model errors, they are usually affected by other sources of errors, e.g., approximations in the observation operator or in the \ac{DA} method itself~\cite{dee-2005, carrassi-2011}. An illustration is provided in Figure~\ref{fig:mec-snap}, where a two-dimensional quasi-geostrophic model is corrected using a neural network. 
The neural network is trained using the analysis increments over one or two days. The discrepancy between the neural network predictions and the actual model error illustrates the difference between model error and analysis increments. 

\begin{figure*}
    \centering
    \includegraphics{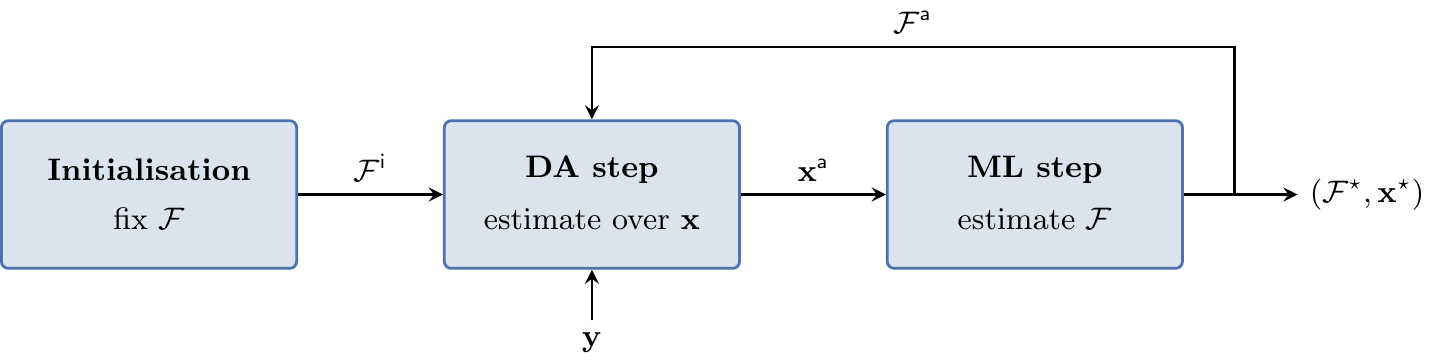}
    \caption{Schematic illustration of the coordinate descent minimisation. \ac{DA} steps are used to estimate the system state from sparse and noisy observations, and \ac{ML} steps are used to estimate the parameters of the statistical correction based on the estimated states. The process can be iterated for increased accuracy in the estimation. Here, $\mathcal{F}^i, \mathcal{F}^a, \mathcal{F}^*$ denote "initial-", "analysis-" and "optimal correction" respectively.}
    \label{fig:daml-loop}
\end{figure*}

More generally, this training process can be interpreted as the first step of a coordinate descent~\cite{bocquet2020bayesian}, where \ac{DA} steps alternate with \ac{ML} steps to learn both the system state and the statistical correction from observations as illustrated in Figure~\ref{fig:daml-loop}. 
By construction, this is an offline learning strategy, because the \ac{ML} step does not start until the entire analysis trajectory is available. This means that one can benefit from all the tools developed for deep learning while relying on the \ac{DA} infrastructure that could be in place and under continuous development over the year (as in, e.g., weather forecast centres). Online learning methods have been recently developed as an alternative~\cite{bocquet-2021, farchi-2021a, malartic-2022, farchi-2022a}. The idea is to use augmented state \ac{DA} techniques to estimate at the same time the system state and the parameters of the statistical correction. Compared to offline learning, online learning approaches naturally fit a sequential context (where observations become available over time) and allow model error correction once the first observation is available. From a machine learning perspective, this two-step approach can also be put in a fully differentiable framework using auto-differentiable to access the gradient of the \ac{DA} itself~\cite{chen2022autodifferentiable}.

\subsection{ML and DA for parameter estimations}
\label{sec: III-B}

Parameters in dynamical systems are considered as additional variables, other than state vector, which determine the dynamic characteristics \cite{shogren2013encyclopedia}.
Online parameter estimation for high-dimensional dynamical systems has been a long-standing challenge~\cite{modares2010parameter}.
In geoscience, hybrid mechanistic-empirical models are widely used for large-scale problems, for instance, in climate~\cite{brown2013ecological} and wildfire~\cite{albini1976estimating} forecasting.
The parameter estimation of these models often relies on case-by-case 
tuning~\cite{lautenberger2013wildland} or posterior diagnosis/analysis~\cite{alessandri2021parameter} that can be computationally difficult due to the complexity 
of the predictive model. Much effort has been given in applying \ac{DA}, especially variational assimilation, for parameter estimation using real-time observations~\cite{smith2009variational,wanders2014benefits}. In particular, an augmented state approach that jointly estimates state and parameters has been used~\cite{jazwinski2007stochastic}. In augmented state approach, the parameters are updated through cross-correlations with the observed state. For this method to work well, there needs to be a substantial correlation between observed values and parameters as pointed out by~\cite{RuckstuhlJanjic20}. Two main difficulties are associated with the estimation of model parameters~\cite{kivman_2003}: one is the strong nonlinear coupling between the parameter and model equations and the second is that the parameters are usually between certain value range (for example parameters are positive as a rule).  Due to both these properties, a \ac{DA} method such as the \ac{EnKF} or variational methods that relies on Gaussian assumptions and uses only the first two statistical moments in the analysis step, needs to be modified in order to be able to deal with probability density functions poorly approximated by the normal distribution. The work of~\cite{annan_etal_2005}, ~\cite{emerick2012history}, and \cite{ruiz2013estimating} presented techniques in parameter estimation that have been successfully applied in low-resolution non-chaotic systems. In addition,~\cite{posselt_bishop_2012, posselt_etal_2014} adopted a new algorithm of~\cite{hodyss2011ensemble} to the estimation of cloud microphysical parameters that use higher than second order moments. ~\cite{RuckstuhlJanjic18} compares several of the nonlinear \ac{DA} algorithms for joint state and parameter estimation and shows the benefits of including higher order moments or physical constraints in \ac{DA}. 
The work of \cite{rochoux2014towards} combines a data-driven simulator for forecasting regional wildfire front position and an EnKF for wind and biomass fuel parameters estimation. In this framework, a surrogate model based on a Polynomial Chaos approximation is iteratively adapted to capture the nonlinearities of the forward model and considerably improves the EnKF efficiency. 
However, all these sample-based algorithms, require stochastic models for parameters to ensure continuous updates of parameters based on new data~\cite{RuckstuhlJanjic18}.  As illustrated in~\cite{RuckstuhlJanjic20}, in real world applications, the accuracy of the parameter estimates is quite sensitive to the stochastic model chosen. Thus, significant effort is required to properly tune the stochastic model of the parameters.

 On the other hand, since the parameters of numerical models are not observed, \ac{ML} is not an obvious method of choice for the estimation of the parameters. However, in a hybrid setting with \ac{DA}, \ac{ML} has shown to bring several benefits to this problem as well. 
\cite{nadler2020neural} proposed the use of \ac{RNN} to replace the standard compartmental model in epidemic modelling for COVID-19. Thanks to its efficiency, this approach can incorporate new data to adjust model parameters via \ac{DA} in real-time.
~\cite{nadler2019data} applied similar ideas to analyse cryptocurrency markets. \cite{li2022joint} implemented deep residual neural networks to surrogate the assimilation process thus enhancing model forecasts. The proposed approach managed to handle both parameter and state estimation with sparse and noisy observations~\cite{li2022joint}.
\ac{ML} can also be used to estimate parameters~\cite{legler_janjic_2022} and their uncertainties as an alternative to the augmented state approach. Although purely offline training produces a good, averaged value of the parameters, we are often in a situation where parameter values might differ due to various reasons like season, or weather situation in \ac{NWP} systems. In this case, combing \ac{ML} with \ac{DA} is quite beneficial, since at the times observations are assimilated, the online improvements to \ac{ML} model can be made.~\cite{legler_janjic_2022} goes beyond deterministic point predictions and learns probabilistic neural networks: a deep ensemble of point estimate neural network and \ac{BNN}. After training, these two types of neural networks are incorporated within a \ac{DA} system, where \ac{DA} is used for state estimation and \ac{ML} for parameter estimation. \acp{BNN} are additionally trained online during the \ac{DA} cycle using a realistic number of forecast/analysis ensemble members allowing further improvements to a \ac{ML} model. By including the parameter estimates obtained from the \acp{BNN} in the \ac{DA}cycle, the results show reduced state errors and increased ensemble spread compared to the case without parameter estimation and with unknown parameters. However, even though the \acp{BNN} can accurately estimate the model parameters and their uncertainties, the high computational cost poses an obstacle. Therefore other methods as random forests are currently considered for parameter estimation problems~\cite{ferrat2018classifying}.

Building efficient forward surrogate models is an alternative solution to reduce the computational burden of inverse modellings~\cite{frangos2010surrogate}, including parameter estimation~\cite{cai2021surrogate}.
The very recent work of~\cite{cheng2022parameter} proposed the use of \ac{GLA} (see~\cite{Cheng2022JSC} or our Section~\ref{sec: IV-B}) to integrate real-time observations for model parameter
adjustment that can yield more accurate future predictions. Learning from a high-fidelity mechanistic model, the authors first constructed a \ac{ROM}- and \ac{ML}-based surrogate for predicting wildfire dynamics. The model coefficients related to the fire spread rate are then consistently updated using real-time satellite observations. The same technology has been applied to nuclear reactor physics~\cite{gong2022efficient} with spatially-sparse observations. By construction, this approach can incorporate observations with flexible length time windows where pure \ac{ML} can get into difficulties with unfixed input dimension.

\subsection{Error specification in DA: traditional and ML methods}
\label{sec: III-C}

Typical \ac{DA} methods are tuned by some statistical parameters, especially those associated with the error terms, i.e. $\epsilon_t^x$ the model error, and $\epsilon_t^y$ the observation error. Assuming that those two error terms are Gaussians, the corresponding model and observation error covariance matrices are noted ${\bQ}_{t}$ and ${\bR}_{t}$ (see Section \ref{sec: II-B}). 
They play a crucial role on the quality of the \ac{DA} reconstruction by  controlling the weight given to the forecast and the observations in the \ac{DA} algorithms. This is illustrated in Figure 2 of~\cite{tandeo2020review}.

Several methods have been proposed in the \ac{DA} literature to jointly estimate ${\bQ}_{t}$ (or alternatively the background covariance ${\bB}_{t}$) and ${\bR}_{t}$, they are summarised in~\cite{tandeo2020review}.  All the methods are based on the innovationdifference, noted $\mathbf{d}^{o-b}_t$, between the observation $\by_t$ and the background projected in the observation space $\mathcal{H}_t \left( {\bx}^{b}_t \right)$, i.e.,
\begin{align}
    \mathbf{d}^{o-b}_t = \by_t - \mathcal{H}_t\left( {\bx}^{b}_t \right).
\end{align}

When considering Gaussian errors $\epsilon_t^x$ and $\epsilon_t^y$, the innovation $\mathbf{d}^{o-b}_t$ is also Gaussian, and its covariance matrix is $\mathbf{H}_t {\bB}_{t} \mathbf{H}_t^\top + {\bR}_{t}$. In order to jointly estimate ${\bQ}_{t}$ (or ${\bB}_{t}$) and ${\bR}_{t}$, innovation alone is not enough, and the authors proposed to examine other innovation statistics (e.g., observation minus analysis, noted $\mathbf{d}^{o-a}$) in the observation space: this is called the Desroziers method~\cite{desroziers2005diagnosis}. Alternatively, several works among~\cite{berry2013adaptive}, based on Mehra theory~\cite{mehra1970identification}, had a look at the lag-innovation statistics, the difference between two consecutive innovations. Both Desroziers and Mehra methods use two different innovations to retrieve the two unknown covariances. 
As an example, the Desroziers innovation statistics should verify the following equations:

\begin{align}
{\bR}_{t} & = \mathbb{E} \left[ \mathbf{d}^{o-a}_t \left(\mathbf{d}^{o-b}_t\right)^\top \right],\\
\mathbf{H}_t {\bB}_{t}\mathbf{H}_t^\top + {\bR}_{t} & = \mathbb{E} \left[ \mathbf{d}^{o-b}_t \left(\mathbf{d}^{o-b}_t\right)^\top\right].
\label{eq:desroziers}
\end{align}

Many works based on the principle of maximum likelihood approaches also use the innovation to find the most likely covariances ${\bQ}_{t}$ and ${\bR}_{t}$~\cite{dee1995line}. In DA, those likelihood-based approaches use either the Bayesian or the frequentist framework. In the Bayesian framework, prior distributions are proposed for the shape parameters of the two covariances (typically, noise levels and spatial correlation lengths). Then, two-stage procedures are used to estimate the state of the system and the posterior distribution of those shape parameters~\cite{stroud2018bayesian}. In the frequentist framework, covariance matrices ${\bQ}_{t}$ and ${\bR}_{t}$ (or parametric versions of them) are tuned to maximise the total likelihood of the state-space representation.
This tuning is often achieved by implementing Expectation-Maximisation (EM) algorithms~\cite{shumway1982approach}.
The latter consists of a two-stage procedure, where the state of the system is estimated, and then covariance parameters are updated on an iterative basis~\cite{tandeo2015offline, pulido2016estimation, dreano2017estimating, pulido2018stochastic, cocucci2021model, chau2022comparison}.

Classical error covariance estimation algorithms (e.g.,~\cite{Desroziers01,desroziers2005diagnosis}) often rely on posterior analysis, and require iterative applications of \ac{DA}. This can be computationally difficult for high-dimensional systems. Furthermore, careful attention must be paid when making the initial guess of error covariances, which may crucially impact the algorithm performance~\cite{menard2016error}. Continuous effort sought to enhance error covariance modelling, in particular, with \ac{ML} techniques. \ac{CELLO} was proposed by Vega-Brown et al.~\cite{vega2013cello} to provide a fast prediction of the observation error covariance $\bR_t$ based on a Bayesian non-parametric learning methods. \ac{CELLO} achieved similar results compared to empirically estimated covariances using manually
annotating sensor regimes.
The authors have also shown that the learned
covariances can provide substantial 
enhancement to state estimation accuracy during online filtering. A \ac{CNN}-based approach, named \ac{DICE}, has been developed to learn the measurement error distribution in a supervised manner~\cite{liu2018deep}. Relying on the Gaussian assumption, \ac{KLD} was employed as the loss function to train $\bR_t$. Thanks to the capacity of \acp{CNN} in capturing local spatial patterns, \ac{DICE} considerably outperformed \ac{CELLO} on both simulated and real data. 

\begin{figure}[h!]
\centering
 \subfloat[Offline training]{\includegraphics[width=0.50\textwidth]{{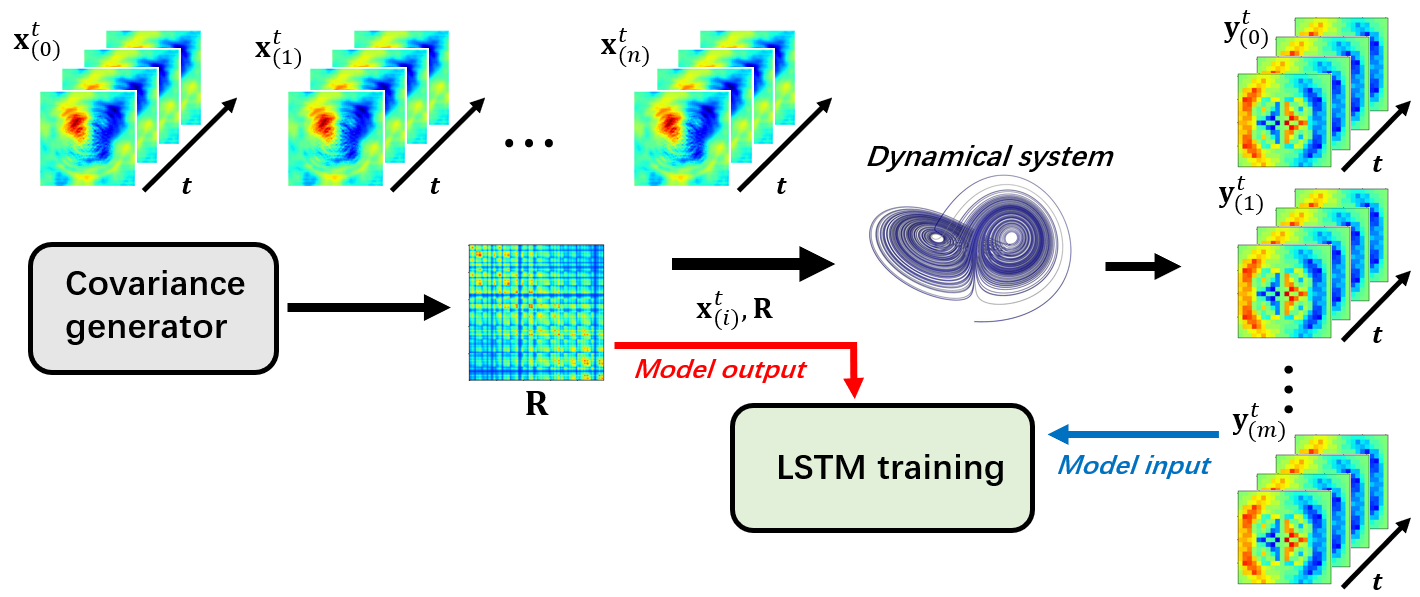}}}\\
 \subfloat[Online prediction]{\includegraphics[width=0.50\textwidth]{{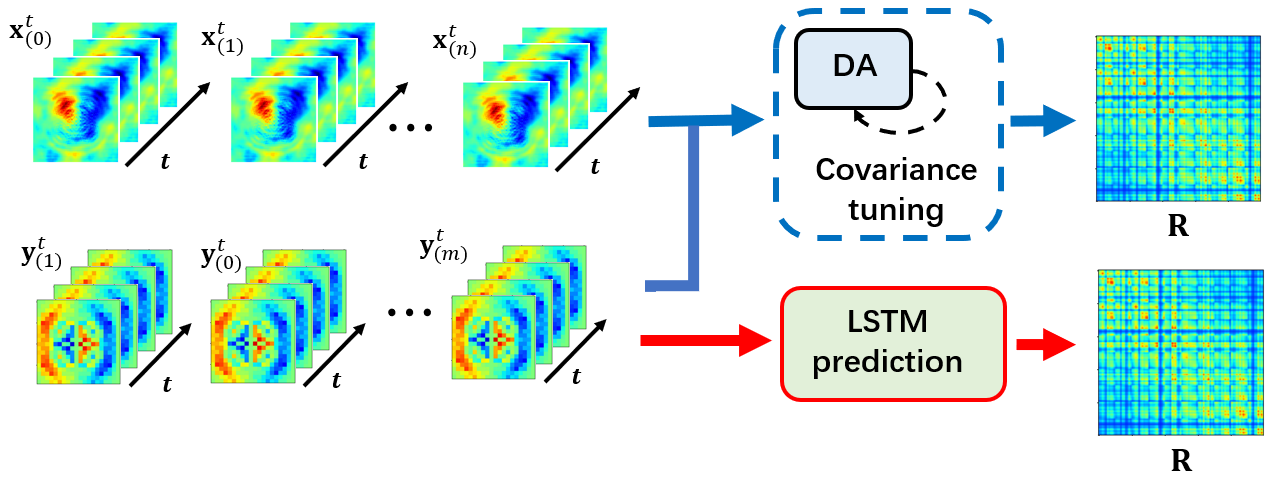}}} 
\caption{Workflow of training and applying \ac{RNN} for observation covariance estimation }\label{fig:covariance_train}
\end{figure}

In fact, $\bR_t$ is often considered time-invariant in a wide range of \ac{DA} applications~\cite{janjic2018representation}.  
Therefore, observed values $\by_t$ at different time steps are jointly considered to predict $\bR_t$. Following this idea, the very recent work of~\cite{cheng2022observation} proposed a \ac{RNN}-based framework for observation matrix specification. More precisely, synthetically generated observation sequences are considered as \ac{RNN} inputs while the corresponding $\bR_t$ is the output target during the training process. In particular, \ac{LSTM} was used to build the model because of its strength in dealing with long-term time dependencies. The workflow of offline model training and online prediction is illustrated in Figure~\ref{fig:covariance_train}. As an important advantage, this approach managed to handle both non-parametric and parametric (e.g., with a pre-selected covariance kernel) covariance modellings. The comparison of different error covariance specification approaches is given in Table~\ref{table:cov}, where the computational efficiency refers to low online computational cost and the temporal dependency signifies if the method can use time-varying data to estimate error covariances.

\begin{table}[h]
\begin{center}

\caption{Comparison of covariance specification approaches}%
\label{table:cov}
\begin{tabular}{ccccc}
\hline
\hline
Methods  &  \makecell{Training \\ free}  & \makecell{good\\ interpretability} & \makecell{computational \\ efficiency} & \makecell{temporal \\ dependency}\\
\hline
\hline
EM~\cite{dreano2017estimating}   & \checkmark   & \checkmark  & \xmark & \checkmark  \\
Desroziers~\cite{desroziers2005diagnosis}  & \checkmark   & \checkmark  & \xmark & \xmark  \\
CELLO~\cite{vega2013cello} & \xmark   & \xmark & \checkmark & \xmark  \\
DICE~\cite{liu2018deep}   & \xmark   & \xmark  & \checkmark & \xmark \\
LSTM-based~\cite{cheng2022observation} & \xmark    & \xmark  & \checkmark & \checkmark \\
\hline
\hline
\end{tabular}

\end{center}
\end{table}

\subsection{End-to-end learning of DA systems}
\label{sec: III-D}

Instead of using \ac{ML} techniques to address difficulties in \ac{DA} algorithms (e.g., model error correction, parameter estimation and error covariance specification), some recent works focused on building end-to-end learning schemes for the whole \ac{DA} system. 
End-to-end learning~\cite{lecun_deep_2015} naturally arises as an appealing feature of deep learning schemes to address a given inverse problem from raw input data through the combination of elementary neural blocks. The key property here is the differentiability of the elementary blocks which leads to the differentiability of the end-to-end architecture. As such, one can train the latter at once using a supervised or partially-supervised learning strategy. This end-to-end learning strategy has been at the core of the breakthroughs of deep learning in many application fields, including signal processing and computer vision~\cite{lecun_deep_2015}. Over the last decade, the elementary blocks or layers available to design neural architectures have also greatly expanded from initial dense, convolution, pooling, activation and recurrent layers~\cite{lecun_deep_2015} to more complex blocks, including among others attention blocks~\cite{vaswani2017attention},  multi-scale neural architectures~\cite{cicek_3d_2016}, finite-difference and spectral solvers~\cite{long_pde-net_2019}, neural optimizer~\cite{andrychowicz_learning_2016}, and physics-informed neural networks~\cite{raissi2019physics}.

\begin{figure*}[htb]
    \centering
    \begin{tabular}{cc}
    \fbox{\includegraphics[height=5.cm]{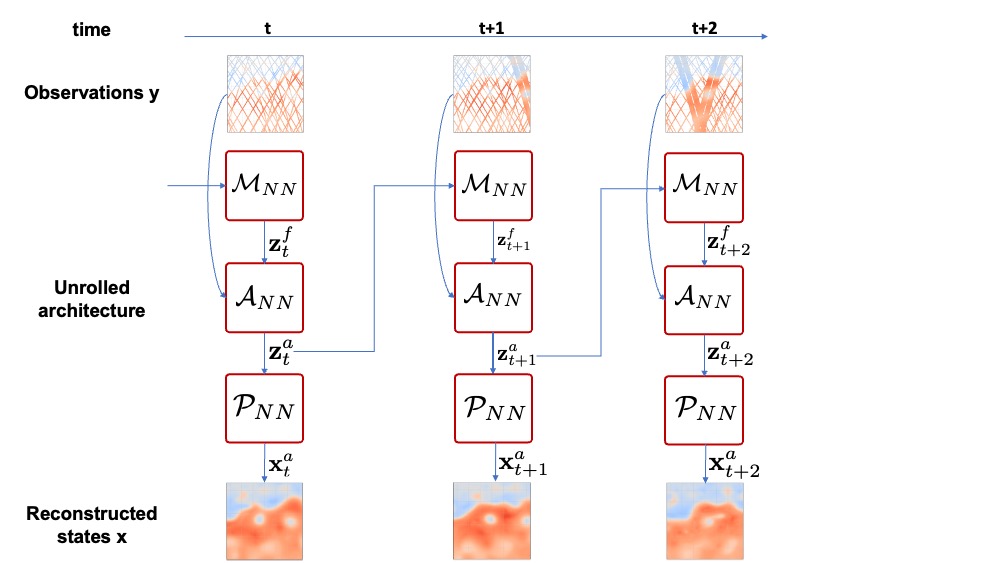}}&
    \fbox{\includegraphics[height=5.cm]{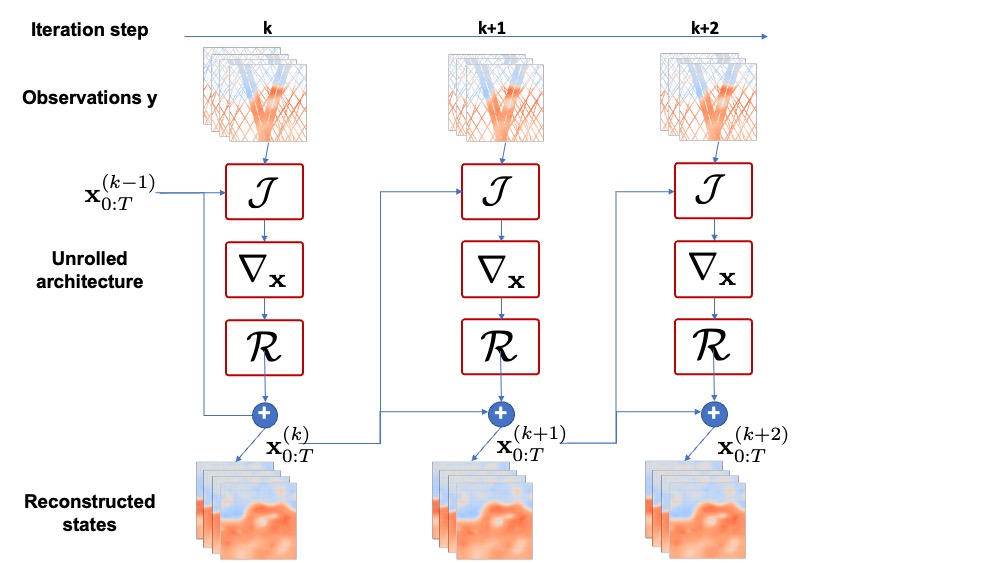}}\\
     {\bf Sequential-DA-inspired scheme}    & {\bf 4DVar-DA-inspired scheme}
    \end{tabular}
    \caption{{\bf Sketch of end-to-end neural architectures for \ac{DA}:} 
    some architectures mimic the forecasting and analysis steps of sequential \ac{DA} at each time step (see Eq.\ref{eq: trainable seqDA}) (left panel) whereas others implement iterative gradient descents for a variational \ac{DA} criterion (right panel).}
    \label{fig:E2E neural DA}
\end{figure*}

This diversity of neural components provides the basis to address \ac{DA} through an end-to-end learning framework. This may simply consist in training a state-of-the-art neural architecture to map observation data to the targeted state sequence or model parameters~\cite{barth_dincae_2020,manucharyan_deep_2021}. Here, we focus on neural approaches which take a closer look at \ac{DA} schemes to design DA-inspired neural schemes. Broadly speaking, as sketched in Fig.\ref{fig:E2E neural DA}, this applies both to sequential \ac{DA} schemes and non-sequential variational \ac{DA} ones:
\begin{itemize}
    \item { Sequential-DA-inspired neural schemes:} in sequential DA, two main operators naturally arise, a forecasting operator to forecast or sample the state at the next time step given the current state and an analysis operator to update the state given new observation data (see Section~\ref{sec: DA principle} for details). From a deep learning point of view, both operators naturally relate to \ac{RNN}. As such, one may explore state-of-the-art \ac{RNN} such as \ac{LSTM} and \ac{GRU}~\cite{lecun_deep_2015} as in~\cite{boudier_dan_2020,revach_kalmannet_2022}. When introducing a latent representation $\mathbf{z}_t$ (for example, obtained from \ac{ROM} as shown in Section~\ref{sec:ROM}) for the physical state $\mathbf{x}_t$ or its \acp{PDF},  this leads to a trainable recursion of the following form at time step $t$:
    \begin{equation}
    \label{eq: trainable seqDA}
        \left \{\begin{array}{ccl}
             \mathbf{z}^{f}_{t},\mathbf{h}_{t}& = & {\mathcal{M}}_{NN} \left ( \mathbf{z}^{a}_{t-1} , \mathbf{h}_{t-1} \right ) \\
             \mathbf{z}^{a}_{t}& = & {\mathcal{A}}_{NN} \left ( \mathbf{z}^{f}_{t}, \mathbf{y}_{t} \right ) \\
             \mathbf{x}^{f}_{t}& = & {\mathcal{P}}_{NN} \left ( \mathbf{z}^{a}_{t} \right ) \\
        \end{array}\right.
    \end{equation}
    where ${\mathcal{M}}_{NN}$, ${\mathcal{A}}_{NN}$ and ${\mathcal{P}}_{NN}$  are neural networks and $\mathbf{h}_{t}$ is the internal state of recurrent networks, if any. One may also explore physically-constrained parameterisations, typically neural \ac{ODE}/\ac{PDE} schemes for the forecasting operator if the underlying physics~\cite{chen_neural_2018,raissi_deep_2018,long_pde-net_2019} are known and/or explicit Kalman recursion rule under additional linear-Gaussian hypothesis for the posterior and the observation operator~\cite{ouala_neural-network-based_2018,revach_kalmannet_2022}. Regarding the learning step, these approaches may adapt classic stochastic optimisation algorithms~\cite{lecun_deep_2015} with randomised re-initialisation steps of the internal states of the recurrent blocks~\cite{boudier_dan_2020};
       
    \item { 4DVar-DA-inspired neural schemes:} from a neural perspective, variational \ac{DA} combines a variational cost (see Section~\ref{sec: DA principle}) and a gradient-based optimizer using an adjoint method~\cite{hascoet2014}.  Assuming that both the observation and dynamical operators are implemented as neural operators, the automatic differentiation embedded in deep learning schemes makes it convenient to apply a gradient descent with respect to state sequence and/or model parameters, with no need to explicitly code the adjoint operators. Besides, trainable optimizers have also emerged as computationally-efficient solvers for minimisation problems~\cite{andrychowicz_learning_2016,hospedales_meta-learning_2020}. The combination of these two elements as proposed in 4DVarNet approach~\cite{fablet_learning_2021} arises as an appealing solution to learn unknown terms in the variational formulation jointly to computationally-efficient solvers for the \ac{DA} problem. It relies on the  weak-constrained 4DVar cost function ${\mathcal J}$ over a time window $[0,T]$ with state variables $\bx_{0:T}$ and observations $\by_{0:T}$ as defined in Equation~\ref{eq:4dvar} in Section~\ref{sec: DA principle}.
 The associated trainable solver exploits the following update rule from some initial condition $\mathbf{x}_{0:T}^{(0)}$:
    \begin{equation}
         \mathbf{x}_{0:T}^{(k+1)} =  \mathbf{x}_{0:T}^{(k)} + {\mathcal{R}}\left [  \nabla_\mathbf{x} \mathcal{J}\left ( \mathbf{x}_{0:T}^{(k)} , \mathbf{y}_{0:T} \right ) \right ]  \\~\\
    \end{equation}
    where $k$ stands for the iteration index, ${\mathcal{R}}$ a recurrent neural network, typically a LSTM, and $\nabla_\mathbf{x} \mathcal{J}$ for the automatic differentiation of the cost function with respect to state $\mathbf{x}_{0:T}$. We want to reiterate that the definition of the cost function $\mathcal{J}$ depends on the initial background $\bx_0^b$ and the transformation operators $\mathcal{H}_{0:T}$ (see Equation~\ref{eq:4dvar}). 
    Beyond these physics-informed parameterisations of the variational cost, especially using neural differential schemes, one may also explore non-sequential representations of the dynamics~\cite{fablet_learning_2021}.
\end{itemize}
From a theoretical point of view, the learning stage for these neural schemes relates to bi-level optimisation problems~\cite{liu_bilevel_2019}, for instance for 4DVar-inspired schemes given by:
\begin{align}
    &\arg \min_{\mathcal{H}_{0:T},\bx_b^0} \mathcal{L} \left ( \left \{(\mathbf{x}_{0:T}^\mathrm{true})_n,(\widehat{\mathbf{x}}_{0:T})_n \right \}_n\right ) \mbox{,  such that,  } \nonumber\\
    &\forall n,(\widehat{\mathbf{x}}_{0:T})_n = \arg \min_{ \mathbf{x}_{0:T} } 
\mathcal{J} \left ( \mathbf{x}_{0:T} , (\mathbf{y}_{0:T})_n \right ) \end{align}

$\{(\mathbf{x}_{0:T}^\mathrm{true})_n,(\widehat{\mathbf{x}}_{0:T})_n  \}_n $ is the training dataset ($n$ is the sample index) and $\mathcal{L} $ the considered training loss. Here $\mathbf{x}_{0:T}^\mathrm{true}$ denotes the theoretical value of the states which are supposed to be known during the training process. Optimal interpolation~\cite{cressie_statistics_2015} solves such a bi-level formulation for a minimum-variance criterion and a linear-Gaussian state-space. End-to-end \ac{DA} schemes then open new research avenues to explore optimal \ac{DA} schemes and reduce estimation biases for nonlinear and/or non Gaussian systems as illustrated for partially-observed nonlinear dynamics by~\cite{fablet_learning_2021}. This also applies to the shift from general-purpose \ac{DA} pipelines to application-centric ones optimized for specific observing systems, states and/or diagnosis variables. Beyond applications on toy examples, recent demonstrations for the reconstruction of sea surface dynamics from satellite-derived observations~\cite{fablet_multimodal_2022} support the relevance of these schemes to advance the state-of-the-art for real \ac{DA} problems. A key challenge is their application to complex spatial-temporal \ac{DA} problems currently solved by operational \ac{DA} systems in climate simulation, operational oceanography and weather forecast. In such contexts, we may emphasise the great flexibility in terms of state definition and model parameterisation opened by the end-to-end learning framework, including for instance augmented state~\cite{ouala2021learning}, multimodal formulation~\cite{fablet_multimodal_2022} and uncertainty representation~\cite{lafon_uncertainty_2022}.

\section{Machine learning assisted by data assimilation and uncertainty quantification}
\label{sec:ML use DA}
In this section, we discuss how \ac{DA} and \ac{UQ} techniques can be used to enhance \ac{ML} models in dynamical systems regarding both prediction accuracy and interpretability. This consists of uncertainty analysis for \ac{ML} approaches (Section~\ref{sec: IV-A}), latent \ac{DA} methods for correcting \ac{ML} surrogate models (Section~\ref{sec: IV-B}), identification of governing equations using \ac{DA} (Section~\ref{sec: IV-C}) and forecasting partially observed dynamical systems (Section~\ref{sec: IV-D}). 

\subsection{Uncertainty analysis for ML approaches}
\label{sec: IV-A}

Different families of \ac{UQ} methods in \ac{ML}, especially \ac{DL}, have been proposed based on Bayesian frameworks, for example, \ac{BNN}~\cite{jospin2022hands} and \ac{MCD}~\cite{gal2016dropout}. The latter consists of ensembles of \ac{NN} optimization iterates or independently trained \acp{NN} (e.g., deep ensembles: DE~\cite{lakshminarayanan2017simple}).

 
In both \ac{BNN} and \ac{DE} methods, epistemic uncertainty is often estimated by looking at an ensemble of trained models where the sampling approach from the set of possible models varies from method to method. Similar to the process of EnKF (see Section \ref{sec: DA principle}) in \ac{DA},
the spread of predictions obtained from different models is then used as an estimate of epistemic uncertainty.

In particular, \ac{BNN} offers a probabilistic interpretation of \ac{DL} models by inferring distributions over the models’ weights. They place a prior distribution over \ac{NN} weights, which induces a distribution over a parametric set of functions. \acp{BNN} thus offer robustness to over-fitting, uncertainty estimates, and can learn from small datasets~\cite{jospin2022hands}.
The Bayesian framework quickly explained here after plays an important role in the foundation of \ac{BNN} and some of the proposed \ac{UQ} methods. 
~\\

Given a set of paired noisy observations $\mathcal{S}_o=\{\alpha_i,\beta_i\}_{i=1}^N$ and a set $\mathcal{A}$ of user assumptions and preferences (e.g., \ac{NN} architecture or likelihood function), the goal is to construct a conditional distribution $p(\tilde{\bx}|\bx,\mathcal{S}_o,\mathcal{A})$ of the quantity of interest $\tilde{\bx}$ given an input vector $\bx$. For each input, it is assumed that the response contains both a deterministic as well as some additive aleatoric noise. We aim at identifying the parameters $\omega$ of the \ac{NN} mapping function $\mathcal{M}_{\omega}$ that fits the function inputs $\bx$ and outputs $\tilde{\bx}$ with maximum likelihood,
\begin{align}
   \tilde{\bx}=\mathcal{M}_{\omega}(\bx)+\epsilon(\bx), \label{eq:UQ1}
\end{align}
where $\epsilon(\bx) $ represents the sum of the aleatoric noises. In the case of a dynamical system, Equation \ref{eq:UQ1} can be written as
\begin{align}
     \bx_{t+1}=\mathcal{M}_{t}(\bx_t)+\epsilon_t^{\bx},
\end{align}
following the notation of Section \ref{sec: DA principle}.
We  then assume a likelihood function $p(\tilde{\bx}|\bx,\omega)$ with parameters to be inferred from the available data. In order to do so, we construct a model function $\tilde{\bx}_{\omega}(\bx)$ of the parameters  that captures the deterministic part of the response and assume a model the distribution for the noise (e.g. multivariate factorized Gaussian likelihood function~\cite{eltoft2006multivariate}). To obtain the posterior distribution for any new input $\bx$, we must marginalise over the model parameters,
\begin{align}
  p(\tilde{\bx}|\bx,\mathcal{S}_o)=\mathbb{E}_{\omega|\mathcal{S}_o}\left [p(\tilde{\bx}|\bx,\omega) \right ],  
\end{align}

where $p(\omega | \mathcal{S}_o)$ is obtained from Bayes’ formula and therefore requires the likelihood of the data to be evaluated. Obtaining the posterior exactly is computationally and analytically intractable. 
Indeed, characterizing uncertainty over \ac{NN} parameters is challenging due to the high-dimensionality and potential complex dependencies of the weights. Moreover the influence of the prior distribution is difficult to be understood.
To address the this obstacle, approximate inference methods~\cite{yao2019quality,abdar2021review} aim to approximate the posterior by another distribution and/or obtaining samples from the posterior. $M$ is denoted as the sampling size. All methods obtain a set of parameters samples $\{{\hat{\omega_j}}\}_{j=1}^M$ that maybe used to approximate the integration via Monte Carlo (MC) estimation as,
\begin{align}
    p(\tilde{\bx}|\bx,\mathcal{S}_o) \approx \sum_{j=1}^M \tilde{\bx}_{\hat{\omega_j}}(\bx)/M,
\end{align}
 which provides an approximate distribution of the total predictive uncertainty. Under Gaussian assumption of the model likelihood, there exists simple close forms for the mean and standard deviation of the posterior prediction. In this case, the approximate total uncertainty standard deviation is a combination of the aleatoric and the epistemic parts of the total uncertainty, respectively. \\
However, in practice $p(\omega | \mathcal{S}_o)$ may be approximated by variational parameters, i.e., by a parametrized function $q_{\theta}(\omega)$. The aim is to approximate a distribution that is close to the posterior distribution obtained by the model. As such, the \ac{KLD} between those two distributions (i.e., $KL(q_{\theta}(\omega) \| p(\omega | \mathcal{S}_o))$) may be minimized with regard to $\theta$. The \ac{KLD} minimisation is also equivalent to maximising the evidence lower bound (ELBO)~\cite{huang2019novel} with respect to the variational parameters.
This procedure is also known as the variational inference (VI)~\cite{gal_uncertainty_2016}. VI is a  technique which replaces the Bayesian modelling marginalisation with optimisation (i.e., replace the calculation of integrals with that of derivatives) which can considerably reduce the computational cost.

The posterior inference may also  be approximated by various Monte Carlo (MC) based methods such as Markov-chain Monte Carlo (MCMC) techniques, drawing \ac{NN} parameters samples from the posterior by using a Markov chain with the distribution of the parameters given the data as its invariant( i.e., stationary distribution).
In particular, Hamiltonian Monte Carlo (HMC) is a reference sampling algorithm but is  extremely computationally demanding~\cite{brooks_mcmc_2011,cobb_scaling_nodate}. Stochastic gradient MCMC approaches based on Langevin and Hamiltonian dynamics have been proposed to alleviate the computational burden of MCMC algorithms thanks to stochastic approximation to the gradients~\cite{welling_bayesian_nodate,nemeth_stochastic_2021,staber_quantitative_2022}. It is important to emphasize that these approaches do not rely on any prior assumptions about the form of the posterior distribution. 

For Deep Ensembles (DE)~\cite{lakshminarayanan2017simple}, the concept is very simple and one needs only to retrain the same network many times with different weights initialisations. The inherent randomness then provides different samples of the trained network parameters, meaning identifying multiple minimums of the \ac{NN} parameter loss landscape (i.e., different \ac{MAP} estimates). If one optimizes the networks with a \ac{MSE} loss, this provides only a measure of epistemic uncertainty. If one optimizes the data log likelihood, it estimate both aleatoric and epistemic uncertainties. 
Variants such as -- Snapshots Ensembles (SEn) obtain the set of multiple minimums without incurring any additional cost as compared to standard training, thanks to letting the  algorithms to converge to $M$ different local optimums during a single optimization trajectory; or --  Stochastic Weight averaging-Gaussian (SWAG) extends SEn by also fitting a Gaussian distribution to the aforementioned local optimums~\cite{maddox_simple_2019}.

Recently, much attention has been given in addressing the \ac{ML} explainablity using \ac{UQ} techniques~\cite{seuss2021bridging, agarwal2021explainable, zhang2022explainable}. The latter can contribute directly to the counterfactual explanations~\cite{sokol2019counterfactual}, for instance, under which condition the decision has been made and with what degree of freedom~\cite{seuss2021bridging}. Furthermore, \ac{UQ} can provide information about model noises, which is crucial for \ac{DA} algorithms when being applied to dynamical systems (see Section \ref{sec: III-C} for details).

\subsection{ML and DA with ROM}
\label{sec: IV-B}
In this section we present the algorithms and applications which combine \ac{DA} and \ac{ROM} (especially \ac{ML}-based ones) to get benefits from both technologies. 

On the one hand, due to the high-dimensionality and the complexity of the transformation function, implementing \ac{DA} for high-dimensional systems can be computationally challenging. Classical solutions consist of dimensionality reduction via projection-based \ac{ROM}, such as \ac{POD} (see Section \ref{sec:ROM}). \ac{DA} is then performed in the reduced space. The optimal choice of the reduced space dimension has also been extensively investigated \cite{arcucci2019optimal,binev2017data,cheng2021observation, xiao2018parameterised}.
Many recent research efforts sought to address this challenge of efficiency by performing \ac{DA} with \ac{ML}-based \acp{AE} (see Section \ref{sec:ROM}). Such algorithms, known as \ac{LA}, can benefit from the efficiency of \ac{ML} and the accuracy of \ac{DA}. More precisely,~\cite{casas2020} proposes to learn assimilated results using \ac{RNN} in a reduced space to enhance future predictions. Similar ideas can be found in~\cite{brajard2020combining} which introduces an iterative \ac{DA}-\ac{ML} scheme is introduced. However, when applying this algorithm, retraining of the neural networks is required when new observations become available. In past two years, online \ac{LA} has raised significant research attention. For sparse and unstructured data, domain decomposition techniques~\cite{xiao2019domain} can also be used to reduce the problem dimension, for example, via community detection through a connection graph~\cite{cheng2021graph, luo2020highly}. 

\begin{figure}[h!]
\centering
\includegraphics[width=0.50\textwidth]{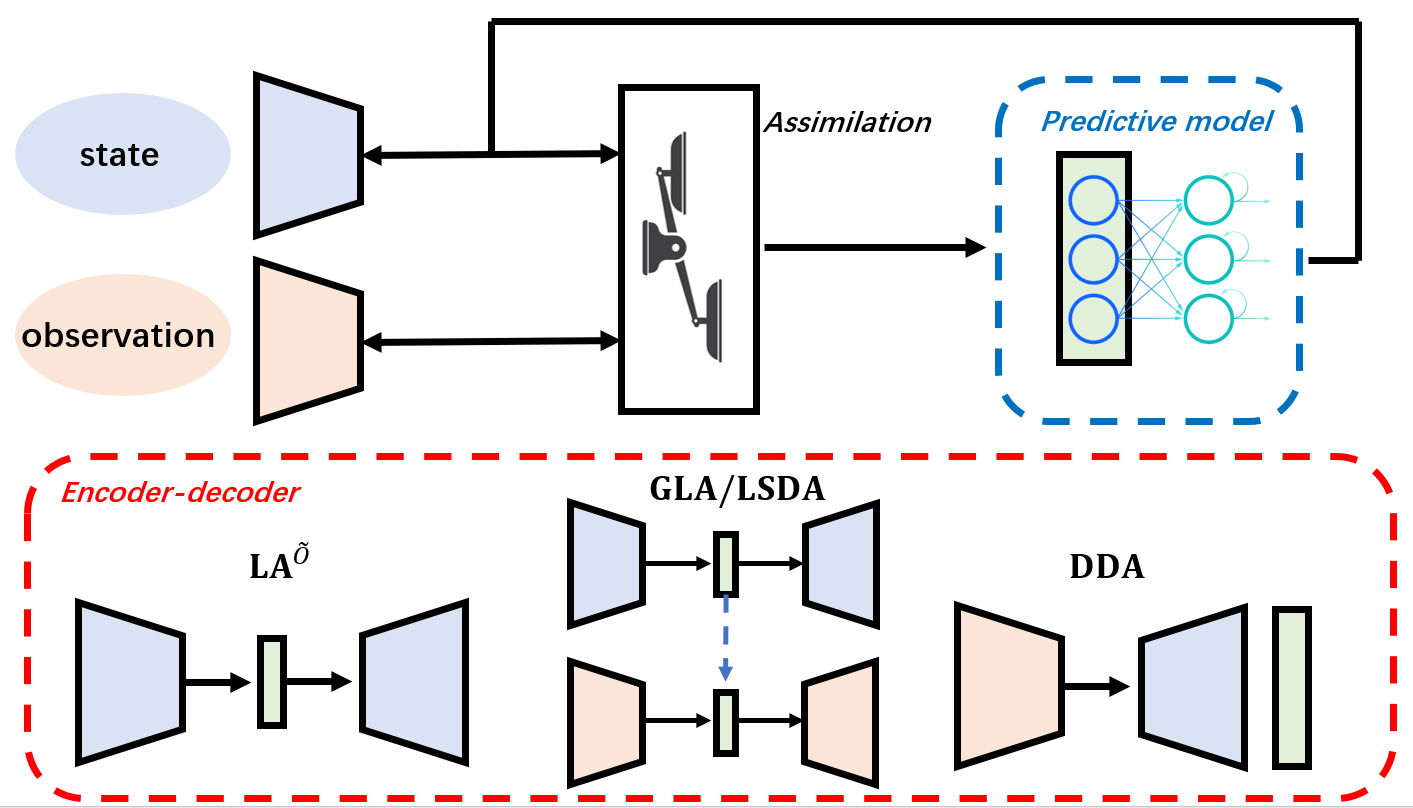}
\caption{Workflow of online \ac{LA} with surrogate modelling and different encoding-decoding strategies}\label{fig:all_LA}
\end{figure}

On the other hand, as discussed in Section~\ref{sec:ROM} and~\ref{sec:predictive}, despite its great efficiency, \ac{ML} surrogate models can introduce prediction errors in a cumulative manner because of the iterative forecasts~\cite{vlachas2018data}. \ac{DA} with \ac{ROM} can address this problem by updating the surrogate prediction consistently using real-time observations collected from local sensors or satellites. 
 As shown in Figure~\ref{fig:all_LA}, this is an online iterative process that can be used to update the starting point of the
next time-level forecast in the latent space, thus improving the accuracy of long-term predictions. Considerable research efforts also sought to adjust the latent error covariance which crucially impacts the assimilation performance (see \cite{tandeo2020review} and Section \ref{sec: III-C}). Ensemble \ac{LA} was introduced in~\cite{zhuang2022ensemble} and~\cite{liu2022enkf} to estimate the background matrix, while~\cite{cheng2022data} employed posterior covariance tuning in the latent space. The proposed online \ac{LA} methods are mainly split into two groups:
\begin{itemize}
    \item $\textrm{LA}^o$~\cite{peyron2021latent,maulik2021efficient}  where observations in the full physical space are used to correct/adjust the reduced-order models;
    \item $\textrm{LA}^{\tilde{o}}$~\cite{amendola2020,mack2020attention,cheng2022data,liu2022enkf} where the state variables and the observations are compressed into a same latent space.
\end{itemize}
The latter can perform more efficient assimilation especially for dense observation mappings, as demonstrated in wildfire forecast~\cite{cheng2022data}, air pollution estimation~\cite{amendola2020} and fluid mechanics~\cite{liu2022enkf}. 
However, encoding state and observation variables into a same reduced space is challenging, especially with highly nonlinear state-observation transformation mappings, which exists in a majority of real-world \ac{DA} problems. 
Therefore, separate \ac{AE} are often required for the states and the observations, leading to heterogeneous latent spaces. 

\begin{table}[h]
\begin{center}

\caption{Comparison of  Latent Assimilation approaches}%
\label{table:LA}
\begin{tabular}{ccccc}
\hline
\hline
Methods  &  \makecell{Reduced \\ state}  & \makecell{Reduced \\ observation} & \makecell{nonlinear \\ mapping} & \makecell{Non-explicit \\ mapping}\\
\hline
\hline
RODDA~\cite{casas2020}   & \checkmark   & \xmark  & \xmark & \xmark  \\
LA~\cite{peyron2021latent,maulik2021efficient}   & \checkmark   & \xmark  & \xmark & \xmark  \\
LA+~\cite{amendola2020,cheng2022data,liu2022enkf} & \checkmark   & \checkmark & \xmark & \xmark  \\
GLA~\cite{Cheng2022JSC}   & \checkmark    & \checkmark  & \checkmark & \xmark \\
LSDA~\cite{mohd2022deep} & \checkmark    & \checkmark  & \checkmark & \xmark \\
DDA~\cite{wang2022deep,storto2021neural} & \xmark   & \xmark  & \checkmark & \checkmark  \\
\hline
\hline
\end{tabular}

\end{center}
\end{table}
To tackle this bottleneck, \ac{GLA} and \ac{LSDA} were proposed in the
very recent works of~\cite{Cheng2022JSC} and~\cite{mohd2022deep} which make use of local surrogate functions (i.e., polynomial functions~\cite{Cheng2022JSC} and \ac{MLP}~\cite{mohd2022deep}) to connect multiple latent spaces. Afterward, variational \ac{DA} can then be performed by solving a local optimisation problem using smooth surrogate functions as shown in Figure~\ref{fig:all_LA}. However, the computation of the local surrogate functions around the predicted latent variables must be performed online, resulting in relatively high computational cost. More importantly, considerable uncertainties can be introduced when mapping the two latent spaces, especially when the choice of the approximation range is inappropriate. Recent research of~\cite{wang2022deep} addresses the difficulty of complex state-observation mapping by proposing a new \ac{DA} scheme, named \ac{DDA}, which trained jointly an observation-domain encoder and a state-domain decoder. A similar idea can be found in the work of~\cite{storto2021neural}. Applying such method, the observation data can be directly transferred to the state space. The characteristics of different \ac{LA} approaches are summarised in Table~\ref{table:LA}.

\subsection{ML for dynamical systems assisted by DA}
\label{sec: IV-C}
\label{se: DD_Identification}
    Except for \ac{ML} surrogate models that learn directly from the state variables (see Section~\ref{sec:predictive} and~\ref{sec: IV-B}), there are several examples of data-driven models derived from observations that show forecasting abilities~\cite{espeholt2022deep, ham2019deep}. Those models can take various forms (e.g. neural networks) but all assume the observations to be perfect and complete, i.e. very little noise and a spatio-temporal complete coverage of the processes of interest. Nevertheless, these conditions are almost never met in reality and dynamical systems of natural processes are usually observed noisily and sparsely. \ac{DA} techniques, on the other hand, can provide a direct methodological formulation that supports the inference of dynamical systems. This formulation can be obtained from a collection of observations that can be irregular both in space and time, noisy, and may also miss some of the degrees of freedom that constitute the underlying dynamics. Following the notation defined in Section~\ref{sec: II-B}, let us consider the following state space model:
\begin{equation}
\left \{
\begin{array}{ccl}
{\mathbf{x}}_{t} &= \mathcal{M}_t(\mathbf{x}_{t-1}, \epsilon^{\mathbf{x}}_{t-1})\\
\mathbf{y}_{t} &= \mathcal{H}_t(\mathbf{x}_{t},\epsilon^{\mathbf{y}}_t)
\label{eq:SSM}
\end{array}\right.
\end{equation}
 The prior errors $\epsilon^{\mathbf{x}}_t$ and $\epsilon^{\mathbf{y}}_t$ are considered as random processes accounting for the uncertainties in the dynamical and observation models.

In an identification scenario, neither the dynamical model $\mathcal{M}_t$ nor the state variable $\mathbf{x}_t$ is known. Instead, we are only provided with observations $\mathbf{y}_t$ that are related in some way to the hidden states through the observation operator $\mathcal{H}_t$. From this point of view, state-of-the-art identification approaches can be discussed based on the nature of the elements of the state space model (Equation \eqref{eq:SSM}).


\subsubsection{Noise-free, direct measurements of the state variables}
When provided with direct measurements of $\mathbf{x}_t$ and assuming that the model and observation noise $\epsilon^{\mathbf{x}}_t$ and $\epsilon^{\mathbf{y}}_t$ are zero, the problem may be regarded as the identification of the most appropriate basis function that explains the temporal variability of the observations. 

One may distinguish data-driven approaches into two prominent families. The first (traditional) category involves an expansion of $\mathcal{M}_{t}$ as a combination of nonlinear basis, where polynomial representations are typical examples~\cite{paduart_identification_2010}. The combination of such representations with sparse optimisation techniques, as shown in \ac{SINDy} framework (see Section~\ref{sec:predictive}), recently opened new research avenues in the context of deriving interpretable dynamical models (see~\cite{brunton2016discovering,rudy2017data} and Section~\ref{sec:predictive}). The \ac{SINDy} methodology has the advantage of interpretability and fewer parameters compared to other \ac{ML} models, which significantly reduces the chances of overfitting. These models were successfully applied to a variety of canonical problems in fluid dynamics~\cite{brunton2016discovering,loiseau2018sparse}, electro-hydrodynamic~\cite{guan2021sparse} and magnetohydrodynamics~\cite{PhysRevE.104.015206}. The main drawback of these approaches remains is that they rely on estimates of the derivatives of the time series. This reliance hampers the direct application to real problems where data can be noisy and irregular. \ac{DA} allows to bypass some of the issues in \ac{SINDy} by providing estimates of state space variables. The work of~\cite{fasel2022ensemble} applied the bootstrap technique in an ensemble-SINDY modelling to address the challenge of noises and uncertainties in observation data.

A second category adopts a \ac{ML} point of view and states the identification issue as a regression problem between consecutive observations. Beyond non-parametric regression models based on analog forecasting~\cite{lorenz1996predictability}, recent state-of-the-art research investigates several methodologies based on different \ac{ML} tools. For instance, reservoir computing approaches~\cite{maass2002real,jaeger2004harnessing} were shown to be well-suited for learning dynamical systems from data~\cite{pathak2017using,patak_PhysRevLett.120.024102}. Furthermore, the link between residual neural networks (ResNets) and \ac{ODE} motivated a large body of work in deriving differential equations that are parameterised by neural networks~\cite{fablet_bilinear_2017,chen2018neural,ouala2021learning}. These new techniques show great flexibility and can be applied to a variety of problems. They can also build from the extensive advances in neural networks and deep learning to tackle challenging issues such as discontinuities in the observations~\cite{herrera2020neural}. These methods, however, may suffer from generalisation issues, which motivate the use of various regularisation techniques, based on prior knowledge of the dynamics, to promote generalisation and interpretability of these data-driven 
models~\cite{raissi2019physics,wang2020incorporating,ma2020combined,wang2020towards,jin2021nsfnets,cai2021physics,kashinath2021physics,mahmoudabadbozchelou2021data,kharazmi2021hp,lucor2022JCP}.


\subsubsection{Noisy observations of the state variables}
When the observation operator $\mathcal{H}_t$ relates to all the states $\mathbf{x}_t$ of the system through an irregular space-time sampling  and the noise processes $\epsilon^{\mathbf{x}}_t$ and $\epsilon^{\mathbf{y}}_t$ are not zero, the derivation of governing equations typically passes through an inversion step. This inversion means that one should estimate the state variables $\mathbf{x}_t$ from the observations in order to perform the identification.
To address this challenge we proposed to leverage on \ac{DA} in a similar fashion to what is described in Section~\ref{sec: III-A} for the estimate of the model error and the construction of a hybrid physics+ML model. Here however, we assume that the model $\mathcal{M}_t$ (see, for example, Equation \eqref{eq:4dvar}) is fully unknown. The lack of a physical model, renders the first cycle of DA/ML very critical: as no model is known at the first cycle, the analysis provided by the \ac{DA} can be very far from the underlying ``truth'' and many optimisation cycles may be required, or in the worst case the procedure could fail to converge. To mitigate this, in some cases, a purely data-driven interpolation (e.g., Kriging) can be performed in place of the first \ac{DA} step to produce the first estimate of $\mathbf{x}_{0:T}$. Another option is to emulate the dynamical system $\mathcal{M}_t$ using analog forecasting methods, and plug it into an ensemble DA technique \cite{lguensat2017analog, zhen2020adaptive}. In absence of any original physical-based model, the data-driven model can only reconstruct dynamics on variables of the system that are observed, even though the problem can be circumvented by using Takens's delay embedding theorem~\cite{takens_theorem,gottwald2021combining} as it is detailed in Section~\ref{sec: IV-D} . If an original model is available, it has been shown that it was possible to correct the dynamics, including non-observed variables~\cite{wikner2021using}. At the end of the DA-ML cycles, the analysis can be used as an initial condition and the data-driven model as a forecast model. 
    
The above approach has shown to improve the forecast skill of small-dimensional dynamical systems~\cite{brajard2020combining} and to be equivalent to an expectation maximisation method~\cite{bocquet2020bayesian}. 
     In~\cite{brajard2020combining}, it is also shown that the data-driven model trained on noisy and sparse data has a skill of the same order of magnitude as a data-driven model trained on complete and noiseless observations. In the cases where the dynamical system is partially known, fewer \ac{DA}/\ac{ML} cycles are necessary~\cite{farchi2021using, brajard2021combining}. Sensitivities studies using \ac{UQ} (see Section~\ref{sec: IV-A}) showed that this approach was not very sensitive to the density of observations, up to a certain point. 
     This is likely to be case specific to vary depending on the application. It has also been shown that the approach is correcting the effect of noisy observation, but the final result is still very sensitive to the noise in the data, as confirmed by other studies~\cite{sangiorgio2021forecasting}.
    Implicitly, all studies on a data-driven models trained on reanalysis~\cite{rasp2020weatherbench, kurth2022fourcastnet} are doing one cycle of this method: first \ac{DA} to produce a reanalysis and then \ac{ML} to train a data-driven model. One obvious limitation of this approach is the computing cost. Several iterations with successive application of the \ac{DA} and the training of the data-driven model are necessary. Therefore, the question of finding the compromise between the improvement of the model and the corresponding cost is crucial.
    Another open question about the use of data-driven or hybrid models in forecast experiments is whether the improvement brought in forecast skill can help understand the deficiency of the current physical-based model. In that perspective, explainable \ac{AI} tools~\cite{mcgovern2019making, toms2020physically, irrgang2021towards} are needed to build operational and trust-worthy systems.

\subsection{ML with DA for partially observed dynamical systems}
\label{sec: IV-D}
In practice, high-dimensional dynamical systems are often only partially observable~\cite{ayed2019learning,boots2011online}.
Let us consider the same problem as Section~\ref{sec: IV-C} with observations only related to a subset of the state vector $\mathbf{x}_t$. Therefore, the derivation of meaningful (Markovian) governing equations in the observation space is (as long as the governing equation of $\mathbf{x}_t$ cannot be decoupled) not possible. This issue was discussed in depth by~\cite{levine2022framework} in the context of closure modelling of a known mechanistic or empirical model. In their work,~\cite{levine2022framework} constructed a
mathematical framework that unifies many of the common approaches for blending mechanistic and \ac{ML} models. The authors studied both discrete and continuous time models and discussed \ac{ML} based closure models that can be both memoryless (Markovian) and memory-dependent. Their representations were also combined with \ac{DA} methods to mitigate noise. 

When there is no prior knowledge about the dynamics, a popular path is based on the phase-space reconstruction methodology. In this framework, we seek at projecting the observations into a higher dimensional space that forms an embedding of the hidden state space variables $\mathbf{x}_t$. The temporal evolution of the variables of the embedding is then deterministic and can, in theory, be used to define a model. The most employed embedding methodology in signal processing is the celebrated Takens delay embedding theorem~\cite{takens_theorem}. It shows that by considering delayed observations, one can unfold a phase space that can be topologically similar to the one of the unseen state variables. Several identification techniques have been used on such representations, including polynomial representations~\cite{Abarbanel1996Modeling_chaos}, recurrent neural networks~\cite{NN_takens}, support vector regression~\cite{SVM_takens}, non-parametric models~\cite{berry2016forecasting} and reservoir computing~\cite{gauthier2021next}. Delay embedding representations were also combined with \ac{DA} frameworks in~\cite{gottwald2021combining} to infer dynamical models from noisy and partial observations.

Interestingly, the idea of using delay embeddings of the observations can also be found at the heart of recent advances in the inference of latent spaces in state space models based on deep learning architectures~\cite{he_deep_2015,krishnan_structured_2016,chen2018neural,nguyen2019emlike}. In such methodologies, latent variables are inferred from a posterior distribution given a sequence of observations. This posterior distribution is parameterised by a neural network and optimized using the evidence lower bound. These frameworks have the advantage of bypassing classical assumptions used in traditional \ac{DA} algorithms such as the Gaussianity of the noise, but suffer, similarly to all models defined based on delay representations, from the problem of correctly parameterising the embedding parameters. 

The parameterisation of a delay embedding~\cite{takens-params-1,Takens-params-2} is a complex task (especially when given high dimensional observations) and the model-making is highly sensitive to this parameterisation. To address these limitations, the Neural Embedding of Dynamical Systems framework (NbedDyn)~\cite{ouala2020learning,ouala2022bounded} proposed to solve the embedding problem jointly with the optimisation of a dynamical model. Specifically, NbedDyn defines a new latent state $\mathbf{z}_t$ as follows:
\begin{align}
{\mathbf{z}_t}^T = [\mathcal{R}(\mathbf{x}_{t})^T, \mathbf{y}_{t}^T]
\label{eq:aug_vect}
\end{align}
with $\mathbf{y}_{t} $ the unobserved component of latent state $\mathbf{z}_t$ and $\mathcal{R}$ an invertible operator used to reduce the dimensionality of the observations. The augmented latent space evolves in time according to the following state space model:
\begin{equation}
\left\{
\begin{aligned}
&{{\mathbf{z}}}_{t} = \mathcal{M}_{\theta,t}(\mathbf{z}_{t-1}, \epsilon^{\mathbf{z}}_{t-1})\\
&{\mathbf{y}}_{t} = \mathcal{R}^{-1}(\mathbf{G}{\mathbf{z}_t}) + \epsilon^{\mathbf{y}}_t
\label{eq:aug_ode}
\end{aligned}
\right.
\end{equation}
where $\mathcal{M}_{\theta,t}$ is the approximate dynamical operator of $\mathcal{M}_t$ and $\mathbf{G}$ is a projection matrix that satisfies $\mathcal{R}(\mathbf{x}_{t}) = \mathbf{G}\mathbf{z}_{t}$. The optimisation of the parameters of the model $\mathcal{M}_{\theta,t}$, as well as the reconstruction of $\mathbf{y}_t$, are carried out jointly using \ac{DA}. Several optimisation strategies can be defined, depending on the form of the dynamical and observation model as well as the uncertainties $\epsilon^{\mathbf{z}}_{t}$ and $\epsilon^{\mathbf{y}}_{t}$. For instance, when considering noise-free observations, a 4D-var formulation was used in~\cite{ouala2020learning,ouala2022bounded} to derive nonlinear dynamical models from partial observations of the state space. In related works, a \ac{KF}-based identification was proposed for linear dynamical and observation models with Gaussian uncertainties~\cite{10.1007/978-3-031-18988-3_13, egusphere-2022-1316}.

In practice, when using phase space reconstruction techniques, one should not forget about the assumptions that this theory is built on. For any embedding to work, we are assuming that the dynamical model in Equation \eqref{eq:SSM} exists and can be represented by an ordinary differential equation~\cite{Sauer1991}. For several realistic applications, this \ac{ODE} may not exist or can have an extremely large dimension. In geosciences, for instance, the dimension of a state space variable can reach $O(10^9)$~\cite{carrassi2018data}. In these situations, reconstructing such high-dimensional phase space becomes significantly more challenging. Reducing the dimension of the problem as demonstrated in~\cite{ouala2020learning,ouala2022bounded} can help making this problem tractable but may lead to closure issues. In practice, the model returned by any embedding technique can be complemented by an appropriate closure. The form of this closure term can be deterministic using, for example, the framework of~\cite{levine2022framework} or stochastic through an appropriate calibration of a noise forcing.
Dynamical system identification approaches introduced in Section~\ref{sec: IV-C} and~\ref{sec: IV-D} are summarised in Table~\ref{table:gove_eq}, where the partial observation stands for partial observations of the state space and the computational efficiency refers to low online computational cost.

\begin{table*}[h]
\begin{center}

\caption{Comparison of governing equations identification approaches}%
\label{table:gove_eq}
\begin{tabular}{ccccc}
\hline
\hline
Methods  &  \makecell{good \\ interpretability}  & \makecell{noisy \& irregular \\ observation} & \makecell{partial \\ observation} & \makecell{computational \\ efficiency}\\
\hline
\hline
polynomial~~\cite{paduart_identification_2010}   & \checkmark   & \xmark  & \xmark & \checkmark   \\
SINDY~\cite{brunton2016discovering,rudy2017data} & \checkmark   & \xmark  & \xmark & \checkmark  \\
ResNet-based ~\cite{fablet_bilinear_2017,chen2018neural,ouala2021learning} & \xmark   & \xmark & \xmark & \checkmark  \\
DA-ML cycle~\cite{brajard2020combining}   & \xmark   & \checkmark  & \xmark & \xmark \\
delay embedding~\cite{he_deep_2015,krishnan_structured_2016,chen2018neural,nguyen2019emlike} & \xmark    & \xmark  & \checkmark & \xmark \\
\hline
\hline
\end{tabular}

\end{center}
\end{table*}

\section{Other approaches, challenges \&  perspectives}
\label{sec: V}
In this section, we briefly introduced some other approaches and future works combining \ac{ML} with \ac{DA}/\ac{UQ} that have not been discussed in detail.

\paragraph{Forecasting Multi-scale dynamical systems}

The question of defining multi-scale representations of dynamical systems is an established practice in theory-guided modelling and empirical representations of dynamical systems. In chemical dynamics for instance, several types of reactions are described by stiff differential equations~\cite{noyes1974oscillatory}. In finances, stochastic representations are a common tool for representing the impact of unresolved parameters on large-scale quantities such as stock prices~\cite{sharp1990stochastic}. In geosciences, The continuity of phenomena across spatiotemporal scales motivated a large body of work for the derivation of multi-scale models that can be both deterministic~\cite{piomelli1999large} and stochastic~\cite{holm2015variational,memin2014fluid,chapron2018large}. From a \ac{ML} of view, defining multi-scale dynamical systems from data has been investigated mainly in prototypical scenarios. The work of~\cite{champion2019discovery} developed sampling strategies for the definition of multi-scale models using state space observations and the definition of closure models for approximating the impact of small-scale variables in the resolution of \ac{PDE} models is by today standards a common practice~\cite{frezat2021physical,frezat2022posteriori,vinuesa2022enhancing,ouyang2022hybrid}. Rethinking such \ac{ML} solutions in terms of real observations may require considering carefully designed, \ac{DA} schemes, in order to deal with noise and irregularities in the observations.

\paragraph{Mode-switching dynamics} Beyond multi-scale variability, real-world dynamical systems are constantly prone to switching between different dynamical modes. Making data-driven representations aware of such issues may help us understand and predict these tipping phenomena in complex dynamical systems from data. From this viewpoint, recent state-of-the-art works started investigating the possibility of finding canonical bifurcations of dynamical systems from toy examples~\cite{bury2021deep}. Generalising such approaches to real data is a challenging question that requires finding these critical transitions and accounting for their dynamics and aftereffects in real-time with \ac{DA}.  

\paragraph{Learning state-observation mapping in data assimilation}
In operational \ac{DA}, the transformation operator $\mathcal{H}_t$ which maps the state variables to the observations $\by_t$ can be complex and highly nonlinear~\cite{van2010nonlinear,reichle2008data}, leading to difficulties in minimising the cost function (see Equation \eqref{eq:4dvar}). Furthermore, as pointed by~\cite{zou1992incomplete,wang2022deep} and our Section~\ref{sec: IV-D}, the observations are often incomplete and only relate to a subset of state variables. Recently, much effort~\cite{wang2022deep} has been given to compute machine learned transformation operators that can decrease the computational burden, and address the missing information in the observation field.~\cite{liang2023machine} applied fully connected neural networks to surrogate the mapping from brightness temperature to microwave radiometer observations. The learned mapping function was then used as the transformation operator in an \ac{EnKF} for \ac{DA}. Similar ideas can be found in~\cite{jing2019data,storto2021neural,Cheng2022JSC,mohd2022deep}. As mentioned in Section~\ref{sec: IV-B},~\cite{Cheng2022JSC,mohd2022deep} compute the surrogate operator in some reduced latent space can further enhance the computational efficiency. On the other hand,~\cite{pmlr-v139-frerix21a} aimed to learn directly the inverse (i.e., observation-to-state) transformation operator to speed-up the convergence of \ac{DA} algorithms. However, since the inverse mapping in \ac{DA} is often not well-defined, observations are still required during the assimilation process. The idea of learning state-observation mapping is naturally related to some cutting-edge \ac{ML} concepts, such as transfer learning~\cite{zhuang2020comprehensive} and domain adaption~\cite{kouw2019review}. It opens promising avenues in solving multi-domain and multi-physics problems.

\section{Conclusion}
\label{sec:conclusion}
The combination of \ac{ML} with \ac{DA} and \ac{UQ} techniques advanced the state-of-the-art of data-driven modelling in various fields and applications.
In this overview, we presented an (as much as possible to our knowledge) exhaustive description
and discussion of state-of-the-art approaches that involve \ac{DA} (or \ac{UQ}) and \ac{ML}. In particular, we insist on that these hybrid models provide strengths in interpretability and noise reduction. The development trends and future challenges of this fast-growing field are also investigated. Significant space for further breakthrough advances still exists, especially in applying these approaches in operational contexts. 
From a methodological perspective, future research efforts could concentrate on, in particular, the integration of \ac{ML} and \ac{DA} in high dimensional, multimodal and multi-scale systems, such as \ac{NWP} and ocean dynamics \cite{schneider2022esa}.
We hope that this review paper will benefit scientists working in this vibrant area by providing guidance on the use of \ac{DA} and \ac{UQ} in \ac{ML} and vice versa, as well as to prompt further developments.

\section*{Acknowledgement}
SC acknowledges the support of Leverhulme Centre for Wildfires, Environment and Society through the Leverhulme Trust, grant no.  RC-2018-023. SC, CQ and RA acknowledge the support of the PREMIERE project (grant no. EP/T000414/1). CQ and RA acknowledge the EPSRC grant EP/T003189/1 Health assessment across biological length scales for personal pollution exposure and its mitigation (INHALE), and the EPSRC grant EP/V040235/1 New Generation Modelling Suite for the Survivability of Wave Energy Convertors in Marine Environments (WavE-Suite). DX would like to acknowledge the support of EPSRC grant: PURIFY ($EP/V000756/1$) and the Fundamental Research Funds for the Central Universities.
MB, JB, AC and AF acknowledge the support of the project SASIP (grant no.~353) funded by Schmidt Futures – a philanthropic initiative that seeks to improve societal outcomes through the development of emerging science and technologies. TJ would like to thank DFG for her Heisenberg Programm Award (JA 1077/4-1). WD would like to acknowledge the National Natural Science Foundation of China under Grant 61976120 and the Natural Science Key Foundation of Jiangsu Education Department under Grant 21KJA510004.

\section*{Acronyms}
\begin{acronym}[htp]
\footnotesize{
\acro{NN}{Neural Network}
\acro{UQ}{Uncertainty quantification}
\acro{ML}{Machine Learning}
\acro{LA}{Latent Assimilation}
\acro{DA}{Data Assimilation}
\acro{AE}{Autoencoder}
\acro{VAE}{Variational Autoencoder}
\acro{CAE}{Convolutional Autoencoder}
\acro{VAE}{Variational Autoencoder}
\acro{GNN}{Graph Neural Network}
\acro{BNN}{Bayesian Neural Network}
\acro{BLUE}{Best Linear Unbiased Estimator}
\acro{RNN}{Recurrent Neural Network}
\acro{CNN}{Convolutional Neural Network}
\acro{LSTM}{Long Short-Term Memory}
\acro{POD}{Proper Orthogonal Decomposition}
\acro{PCA}{Principal Component Analysis}
\acro{MAE}{Masked Autoencoders}
\acro{NLP}{Natural Language Processing}
\acro{PGD}{Proper Generalized Decomposition}
\acro{ROM}{Reduced-Order Modelling}
\acro{DMD}{Dynamic Mode Decomposition}
\acro{CFD}{Computational Fluid Dynamics}
\acro{PDF}{Probability Density Function}
\acro{NWP}{Numerical Weather Prediction}
\acro{MSE}{Mean Square Error}
\acro{MAE}{Mean Absolute Error}
\acro{ODE}{Ordinary Differential Equation}
\acro{PDE}{Partial Differential Equation}
\acro{AI}{Artificial Intelligence}
\acro{DL}{Deep Learning}
\acro{KNN}{K-Nearest Neighbours}
\acro{RF}{Random Forest}
\acro{KF}{Kalman Filter}
\acro{EKF}{Extended Kalman Filter}
\acro{GLA}{Generalised Latent Assimilation}
\acro{3Dvar}{Three-dimensional variational data assimilation}
\acro{4Dvar}{Four-dimensional variational data assimilation}
\acro{MLP}{Multi layer percepton}
\acro{DDA}{Deep data assimilation}
\acro{LSDA}{Latent Space data assimilation}
\acro{CELLO}{Covariance Estimation and
Learning through Likelihood Optimization}
\acro{DICE}{Deep Inference for Covariance} \acro{KLD}{Kullback–Leibler Divergence}
\acro{VarDA}{Variational data assimilation}
\acro{EnKF}{Ensemble Kalman Filter}
\acro{KF}{Kalman Filter}
\acro{GRU}{Gated Recurrent Unit}
\acro{SINDy}{Sparse Identification of Nonlinear Dynamics from Data}
}
\acro{MCD}{Monte-Carlo Dropout}
\acro{DE}{Deep Ensembles}
\acro{BNN}{Bayesian Neural Network}
\acro{MAP}{Maximum a Posteriori}
\end{acronym}

\footnotesize{
\bibliographystyle{IEEEtran}  
\bibliography{main}}



\section*{Authors bibliography}
\textbf{Sibo Cheng} (sibo.cheng@imperial.ac.uk)
completed his Ph.D at Universite Paris-Saclay, France, in 2020. He is currently a research associate at the Data science institute, Department of Computing of Imperial College London. His research mainly focuses on physics-related machine learning, data assimilation and reduced order modelling for high-dimensional dynamical systems with a wide range of applications including wildfire prediction, computational fluid dynamics, microfluidic drop modelling and Medical AI\\

\textbf{César Quilodrán-Casas} (c.quilodran@imperial.ac.uk)
' areas of expertise are data-driven methods and machine learning. He was awarded his PhD from Imperial College London on data-driven oceanography in the Space and Atmospheric Physics Group. As a Research Associate at the Data Science Institute, Imperial College London he has worked in diverse multidisciplinary teams and applied his ML expertise on adversarial and generative networks to different applications such as urban air pollution, microfluidics, and wave energy.  \\

\textbf{Said Ouala} (said.ouala@imt-atlantique.fr) received the M.S. degree in AI and signal processing from the Sorbonne University, Paris, France, and the Ph.D. degree in AI for geophysical dynamics from IMT-Atlantique, Brest, France. He is currently a postdoctoral research assistant in the Stochastic Transport in Upper Ocean Dynamics (STUOD) program. His research interests are at the intersection of dynamical systems and artificial intelligence, with applications in geosciences.\\

\textbf{Alban Farchi} (alban.farchi@enpc.fr) recieved his PhD in physics and environmental sciences (Uni. Paris-Est, 2019). He is a recently hired permanent researcher at CEREA. He works in the field of data assimilation for the geosciences with application to atmospheric chemistry. Since 2020, he has been working with ECMWF on the use of machine learning techniques to correct model error in data assimilation and forecast applications.\\

\textbf{Che Liu} (che.liu21@imperial.ac.uk)  is currently pursuing the Ph.D. degree with the School of Earth Science and Engineering, Imperial College London, London, UK. He is also affiliated with Data Science Institute, Imperial College London, London, UK. His research is at the interface between deep learning and multimode medical data processing, such as biomedical signal processing, medical image segmentation, clinical text processing, and image-text fusion. \\

\textbf{Pierre Tandeo} (pierre.tandeo@imt-atlantique.fr) was born in France in 1983. He received the M.S. degree in applied statistics from Agrocampus Ouest, Rennes, France, and the Ph.D. degree from the Oceanography from Space Laboratory at IFREMER, Brest, France, in 2010. Then, he spent two years as a Postdoctoral Researcher with the Atmospheric Science Research Group, University of Corrientes, Argentina, and three years at Télécom Bretagne, Brest, France. Since 2015, he is an associate professor at IMT Atlantique, Brest, France, and a researcher at Lab-STICC, CNRS, France. Since 2019, he is an associate researcher at the Data Assimilation Research Team, RIKEN Center for Computational Science, Kobe, Japan. His main research interests are focused on IA, data assimilation, and inverse problems for geophysics.\\

\textbf{Ronan Fablet} (ronan.fablet@imt-atlantique.fr) currently holds a full Professor position at IMT Atlantique  and is a research scientist at Lab-STICC in INRIA team Odyssey. With engineer and PhD degrees in Applied Math and Signal Processing, he has a significant experience in interdisciplinary research at the interface between data science and ocean science, especially space oceanography and marine ecology. He currently leads Research Chair OceaniX on Physics-informed AI for Ocean Monitoring and Surveillance. He co-authored more than 200 articles and communications in peer-reviewed conferences and journals. He is also a member of the scientific committees of French and European programs and institutes for his dual ocean-AI expertise. His current research interest includes deep learning for data assimilation, ocean modeling and ocean observation. \\

\textbf{Didier Lucor} (didier.lucor@lisn.upsaclay.fr) is a CNRS research director and deputy director of the Interdisciplinary Laboratory of Numerical Sciences (LISN), part of Paris-Saclay campus in Orsay France. He received his PhD in Applied Mathematics in 2004 from Brown University, USA and he was a postdoctoral fellow in the department of Ocean Engineering at MIT, USA.
 He is the coordinator of the France West Ercoftac pilot centre. His research interests relate to stochastic modelling, computational mechanics and probabilistic scientific computing, with emphasis on physics-informed machine learning, uncertainty quantification and data assimilation and applications ranging from turbulence, environmental flows, energy systems and biomechanics. \\

\textbf{Bertrand Iooss} (bertrand.iooss@edf.fr) obtained a Ph.D thesis in Geostatistics at the Paris School of Mines (1998) and an habilitation thesis in Statistics at Toulouse University (2009). He works as a senior researcher in industrial statistics at Electricité de France (EDF), having managed (2015-2021) a research project named “Uncertainty quantification and machine learning” for electricity production needs, in particular for the nuclear industry. His main research works involve the design, analysis, modeling, uncertainty quantification and validation of computer experiments, related to nuclear safety and environmental problems. He is currently interested in the topics of interpretability and validation of machine learning techniques. \\

\textbf{Julien Brajard} (julien.brajard@nersc.no) is an associate professor at Sorbonne University (Paris, France) since 2009 and a researcher at NERSC (Nansen Environmental and Remote Sensing Center) since 2018. He works at proposing new methodologies at the crossroads between data assimilation and machine learning. His objective is to improve the forecast of key climate variables by correcting model errors using data, in particular from remote sensing. \\

\textbf{Dunhui Xiao} (xiaodunhui@tongji.edu.cn) is a professor at Tongji University (Shanghai, China). He
obtained his PhD from Imperial College London where he did his Post-doc.  His research interests include numerical modelling with a focus on non-intrusive reduced-order modelling of Navier-Stokes equations, fluid-structure interactions, and multiphase flows in porous media. He is also interested in data-driven modelling, data science, physical data combined machine learning and optimisation. He is the PI of a number of grants. He sits on the editorial boards for a number of journals and he is the reviewer for many journals and the EPSRC. \\

\textbf{Tijana Janjic} (tijana.pfander@physik.uni-muenchen.de) is Heisenberg Professor of Data Assimilation at the Mathematical Institute for Machine Learning and Data Science, KU Eichstaett-Ingolstadt, Germany. She received her PhD in Applied Mathematics from University of Maryland and her BSc in Mathematics from the University of Belgrade. To date, her international and interdisciplinary career led to significant contributions to both theory and real-life applications in the field of data assimilation. Her scientific papers span topics in convective scale data assimilation, numerical methods, uncertainty quantification, large-scale linear and nonlinear constrained optimization and data science. She has been serving as associate editor for Journal of Advances in Modeling Earth Systems and QJRMS.  She is also a recipient of prestigious Heisenberg funding of German Science Foundation (DFG). \\

\textbf{Weiping Ding} (ding.wp@ntu.edu.cn) is a Full Professor with the School of Information Science and Technology, Nantong University, Nantong, China. His main research directions involve deep neural networks, multimodal machine learning, and medical images analysis. He received the Ph.D. degree in Computer Science, Nanjing University of Aeronautics and Astronautics, Nanjing, China, in 2013. In 2016, He was a Visiting Scholar at National University of Singapore, Singapore. From 2017 to 2018, he was a Visiting Professor at University of Technology Sydney, Australia. He has published over 200 articles, including over 80 IEEE journal papers. His fifteen authored/co-authored papers have been selected as ESI Highly Cited Papers. He serves as an Associate Editor/Editorial Board member of IEEE Transactions on Neural Networks and Learning Systems, IEEE Transactions on Fuzzy Systems, IEEE/CAA Journal of Automatica Sinica, IEEE Transactions on Intelligent Transportation Systems, Information Fusion, Information Sciences, Neurocomputing, Applied Soft Computing. He is the Leading Guest Editor of Special Issues in several prestigious journals, including IEEE Transactions on Evolutionary Computation, IEEE Transactions on Fuzzy Systems, and Information Fusion. \\

\textbf{Yike Guo} (yikeguo@ust.hk) is the Provost of the Hong Kong University of Science and Technology science December 2022.
 He is also a Professor of Computing Science in the Department of Computing at Imperial College London. He is the founding Director of the Data Science Institute at Imperial College. His main research interests lie in the field of machine learning and large-scale data analysis management. He is a Fellow of the Royal Academy of Engineering (FREng), Member of Academia Europaea (MAE), Fellow of British Computer Society and a Trustee of The Royal Institution of Great Britain. He received his PhD in Computational Logic from Imperial College in 1993. Professor Guo has published over 250 articles, papers and reports. Projects he has contributed to have been internationally recognised. \\

\textbf{Alberto Carrassi} (alberto.carrassi@unibo.it) is trained as a physicist, with a PhD thesis on dynamical systems. He has been always working at the crossroad between applied mathematics and climate. Research breakthroughs include contribution to the development of algorithms known as the assimilation in the unstable subspace unifying dynamical systems and data assimilation or of a novel paradigm for causal inference using data assimilation for detection and attribution of climate change. In recent years, he contributed to pioneer powerful combined methods merging data assimilation and machine learning, making possible using machine learning with the very sparse and noisy climatic dataset. In 2019 he became Professor at the University of Reading (UK) until November 2021 when he became Professor at the University of Bologna (IT). \\

\textbf{Marc Bocquet} (marc.bocquet@enpc.fr)
 is Professor at École des Ponts (France) and deputy director of CEREA laboratory. He works on the methods of data assimilation, machine learning, inverse problems and environmental statistics,
with applications to dynamical systems, atmospheric chemistry and transport, and meteorology. He is a fellow of ECMWF, and editor for the QJRMS, Foundation of Data Science, and Frontiers in Applied Mathematics and Statistics. He has published over 115 peer-reviewed papers. \\

\textbf{Rossella Arcucci} (r.arcucci@imperial.ac.uk)
is an Assistant Professor (Lecturer) in Data Science and Machine Learning at Imperial College London (ICL). She is an elected member of the World Meteorological Organization and the elected speaker of the AI Network of Excellence at ICL where she represents more than 270 academics working on AI. She has been with the Data Science Institute at ICL since 2017 where she has created, and she leads the Data Assimilation and Machine Learning (Data Learning) group. Her work involves developing AI models for climate, health and environmental impact and she also collaborates with the Leonardo Centre on Business for Society at Imperial College Business School. Ph.D. in Computational and Computer Science in February 2012 and she received the acknowledgement of Marie Sklodowska-Curie fellow from European Commission Research Executive Agency in 2017. She is co-investigator of several grants and projects. 
\end{document}